\documentclass{article}

\usepackage{tcolorbox}
\usepackage[final]{corl_2026} %
\usepackage{xspace}
\usepackage{graphicx}
\usepackage{booktabs}
\usepackage{multirow}
\usepackage{float}
\usepackage{placeins}
\usepackage{wrapfig}
\usepackage{caption}
\usepackage{enumitem}
\usepackage{titletoc}
\usepackage{amsmath}

\newcommand{\pmm}[1]{\tiny{$\pm$#1}}
\newcommand{\mo}{M} %
\newcommand{\mr}{\widehat{M}} %

\newcommand{\ours}{GROW$^{2}$\xspace}
\newcommand{\ourbench}{GROW$^{2}$Bench\xspace}
\newcommand{\xhdr}[1]{\noindent\textbf{#1}}

\title{\ours: Grounding Which and Where \\ for
Robot Tool Use}

\author{
\textbf{Yuhong Deng}$^{1^\text{*}}$\quad \textbf{Yuyao Liu}$^{1,2\thanks{Equal contribution. \texttt{yuhongdeng@u.nus.edu}.}}$\quad 
\textbf{David Hsu}$^{1}$\\
$^1$National University of Singapore\quad
$^2$Massachusetts Institute of Technology
}

\begin{document}

\maketitle

\begin{abstract}
Can the robot use a plate to cut a cake if no knife is available? Tool use greatly expands robot capabilities, but to use tools creatively beyond their intended functions, the robot faces the challenge of \textit{open-world affordance grounding}: select an open-category object to act as a tool and localize its specific region of action. To this end, we introduce \ours (GROunding Which and Where), which leverages object parts as a natural abstraction to split the grounding process hierarchically into semantic and geometric levels, thus bypassing the need for data-heavy, end-to-end training. Semantically, \ours harnesses the commonsense reasoning of Vision-Language Models (VLMs) to parse a natural-language task instruction, select a suitable object as the tool, and identify task-relevant parts on the tool and the target object. Geometrically, vision foundation models then ground the selected parts into precise 3D regions from a single RGB-D image. Experiments on established benchmarks show that \ours outperforms state-of-the-art baselines on  affordance prediction benchmarks. Further, it achieves zero-shot generalization over open-category objects and outperforms baselines in both simulated and real-world robot tool use experiments. 

\end{abstract}


\section{Introduction}
\begin{figure}[tb]
    \centering
    \includegraphics[width=\linewidth]{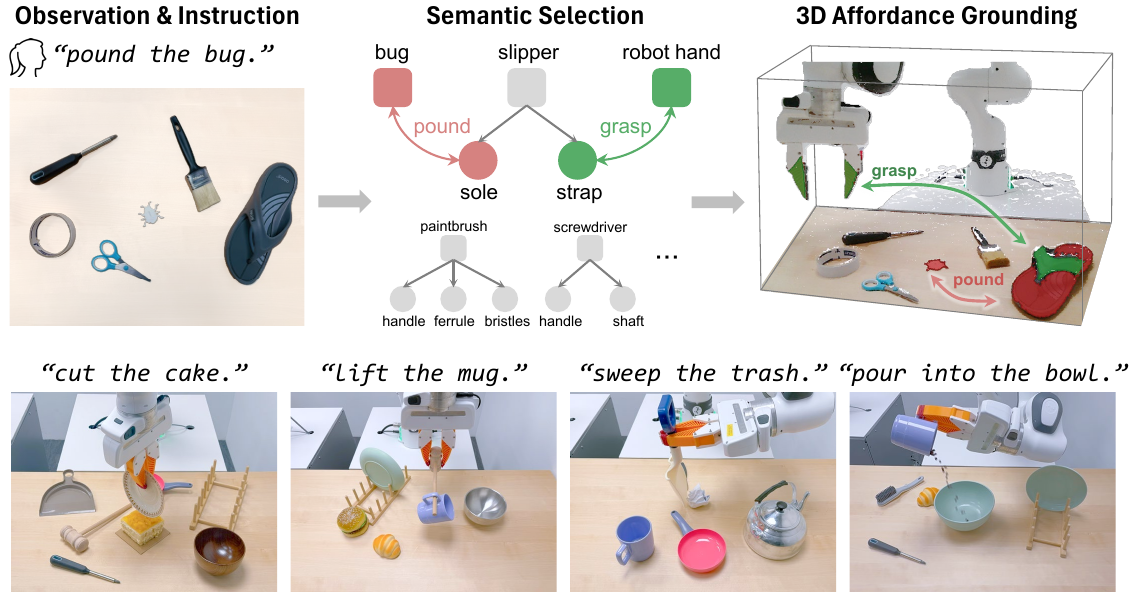}
    \caption{\textbf{\ours (GROunding Which and Where)} decomposes open-world affordance grounding into two levels: (i) semantic selection of an appropriate tool and the task-relevant parts, and (ii) grounding the selected parts into 3D affordance regions. This decomposition enables \ours to leverage a series of foundation models and generalize to open-category objects.}  
     \label{fig:teaser}
\end{figure}

Humans possess a remarkable capacity to use tools in improvised ways: from a cluttered scene of everyday objects, we can identify which object can serve as the needed tool and apply it through the right interaction~\cite{challenges_tooluse, tool_use_survey, improvised_tool} (Fig.~\ref{fig:teaser}). For example, when a knife is unavailable, we may cut a cake with the edge of a plate. We aim to bring this capability to robots, expanding their feasible manipulation strategies and improving robustness in unstructured environments such as homes and warehouses.

To improvise tool use, the robot has to reason about the environment and figure out which object can serve as the tool and where the relevant interaction regions are on that object in order to perform the desired action. Unlike previous works on robot tool use, we do not assume that the tool object is already specified~\cite{task_oriented_grasping, force_tool, roboninja}, or that we have a fixed set of tools with predefined tool-function mappings~\cite{tool_selection_1,robocook,mimicfunc}. Instead, this work addresses the problem of open-world affordance grounding: selecting a tool from a set of open-category objects in the scene and predicting its affordance regions.

There are two main challenges in open-world affordance grounding. (i) The system needs to semantically understand affordances of open-category objects. For example, to cut a cake, a robot should identify an object with a sharp edge, while preferring a plate over a dustpan despite their similar local geometry. (ii) The affordances must be grounded to precise geometric regions for downstream manipulation. This grounding should be conditioned on the task and the environment, and should generalize to open-category objects. Traditionally, people build a large-scale dataset that maps tasks to affordance regions and train an end-to-end neural network~\cite{manipvqa, laso, affordance_2d_1}, which is costly and struggles to generalize to unseen task-object pairs. 

Our key insight is that object parts, such as a blade, rim, and tip, provide a useful abstraction that captures both affordance semantics and the corresponding geometric structure. Using parts as an intermediate representation, we propose \ours (GROunding Which and Where) and decompose affordance grounding into two levels. At the semantic level, \ours leverages the reasoning capabilities of vision-language models (VLMs) to select an appropriate tool and task-relevant parts. At the geometric level, \ours grounds the selected parts into 3D affordance regions using vision foundation models. This decomposition eliminates dependence on large-scale affordance annotations and lets \ours generalize to open-category objects.

Grounding these selected parts in 3D from a single view remains challenging because object parts are often only partially visible or are occluded by other objects or by the object itself. Although multi-view camera setups can alleviate this issue, they are often costly~\cite{voxposer, DistilledField, d3field}. \ours addresses this challenge by combining 2D part grounding with 3D reconstruction. Given a single-view RGB-D observation, it reconstructs a high-fidelity object mesh, registers the mesh to the observed scene, renders the mesh from multiple viewpoints, segments the selected parts in each rendered view, and fuses the resulting masks into complete 3D affordance regions. This enables \ours to recover function-faithful 3D affordances from only a single RGB-D view.

We evaluate \ours on both existing affordance prediction benchmarks and a newly constructed benchmark that introduces the additional challenge of tool selection. \ours outperforms state-of-the-art baselines and generalizes well to open-category objects. We further evaluate the predicted affordances in downstream robot tool-use experiments, including a simulation benchmark with five task types, 85 object assets, and 100 scenes, as well as real-world experiments on a Franka arm. Across these settings, \ours achieves higher task success rates than the baselines, demonstrating that its grounded affordance regions effectively support robot tool use.

\section{Related Work}
\label{sec:relatedwork}

\subsection{Affordance Prediction for Robot Manipulation}
Understanding affordances~\cite{affordance_2, affordance_1, robopoint} is fundamental for robot interaction in unstructured environments. Keypoint-based methods represent affordances as sparse function points~\cite{clasp, k-vil, keypoint_future,gift}, which are efficient for learning and planning but limited in expressiveness, whereas 2D affordance-region methods~\cite{affordance_2d_1, robotABC, affordance_region_2d} integrate well with vision foundation models~\cite{affordance_2d_2,weakly_aff, uad, affordance_vfm} but lack the geometric information needed for manipulation. We instead ground 3D affordance regions. End-to-end 3D affordance prediction methods~\cite{laso, geal} often require substantial 3D training data and generalize poorly to open-category objects. Moreover, most of them assume complete 3D observations~\cite{densematcher}, which is unrealistic. Multi-view fusion methods~\cite{DistilledField, d3field} aggregate 2D affordance predictions across views to leverage foundation models for generalization and to mitigate occlusion. However, the multi-view setting remains inefficient and impractical for real-world deployment. \ours uses part representations to decompose affordance grounding, allowing us to take advantage of different foundation models for generalization. Besides, it can produce complete and function-faithful 3D affordance regions from a single-view observation, without requiring multi-view sensing.

\subsection{Robot Tool Use}
Tool use has been widely studied in robotics and is crucial for enabling robots to extend their capabilities. Prior work equips robots with advanced tool-use skills such as cutting~\cite{roboninja, lbm}, fastening a nut~\cite{nut_fastening}, spreading sauce~\cite{diffusion_policy}, and using chopsticks~\cite{chopsticks}. Despite this progress, most existing approaches assume that the tool and its function are pre-specified and focus primarily on generating actions under constraints~\cite{force_tool, pi05, tool_manipulation_1, toolflownet}. However, in unstructured environments such as homes and warehouses, effective tool use requires the ability to first identify an appropriate object as a tool and then use it to complete the desired task~\cite{challenges_tooluse,creative_llm}. For tool selection, researchers have attempted to harness geometric information to guide the selection process via mathematical analysis~\cite{tool_selection_1, primal_sketch, tool_selection_differentiable} and data-driven approaches~\cite{robocook,tool_Selection_2}. However, these methods often rely on heavy supervision or in-domain data, which limits open-world generalization. In contrast, our method leverages the commonsense knowledge of VLMs to select tools and task-relevant parts in semantic space, and then grounds the selected parts into 3D affordances, supporting effective robot tool use.

\section{Problem Statement}
\label{sec:problem}
We address the problem of open-world affordance grounding in robot tool use. An affordance describes an action possibility arising from the complementarity between an agent and its environment~\cite{affordance_definition}. For robot tool use, we extend this notion beyond robot-object interactions to capture object-object interactions between a tool and a target. We define an affordance as an asymmetric binary relation between two entities that specifies how they interact to realize a particular action. To ground this relation spatially, we localize corresponding interaction regions on the two entities.

For robot tool use, we ground two affordances: the relation between the robot hand and the tool object $o_A$, which enables grasping, and the relation between the tool object $o_A$ and the target object $o_B$, which enables the tool function. The input consists of (i) a single-view RGB-D observation $(I,D)$ containing objects $\{o_i\}_{i=1}^N$ and (ii) a language instruction $l$ describing the task and the target object $o_B$ (e.g., cut the cake). The output consists of the selected tool object $o_A$ and the corresponding grounded 3D interaction regions.

\begin{figure*}[ht]
    \centering
    \includegraphics[width=\linewidth]{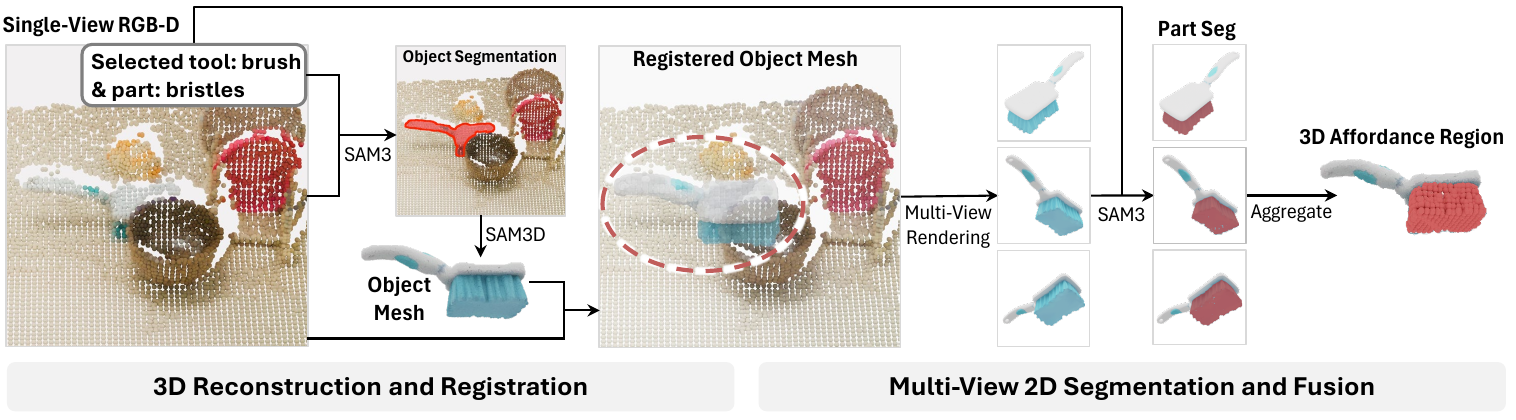}
    \caption{\textbf{3D affordance grounding.} Our method consists of two stages: reconstructing and registering an object mesh from a single-view RGB-D observation and rendering the registered mesh from multiple viewpoints, segmenting the selected parts in each view, and fusing the resulting masks into 3D affordance regions.}
     \label{fig:perception}
\end{figure*}

\section{Method}
\label{sec:method}
We propose \ours (GROunding Which and Where), a novel approach for open-world affordance grounding. The key challenge is the generalization over open-category objects, where end-to-end mapping from tasks to affordances often struggles~\cite{uad, geal}.
To address this, we introduce object parts such as blades, rims, and handles that serve as intermediate cues. Using object parts, we decompose the affordance grounding into two levels (Fig.~\ref{fig:teaser}). Specifically, given a language instruction $l$ and a single-view RGB-D observation $(I,D)$ of the environment, \ours first selects an appropriate tool object and the target object, along with the names of task-relevant parts. It then grounds these selected part names into 3D affordance regions using existing vision foundation models, enabling \ours to generalize across open-category objects.

\subsection{Tool and Part Selection}
\label{subsec:tool_select}
\ours starts with a cluttered scene of multiple objects and a task specified with a language instruction $l$. Instead of finding an appropriate object as the tool solely through geometric features, \ours selects the tool in semantic space with the commonsense knowledge of VLMs. Specifically, \ours first extracts the names of all objects and their parts as candidates. After that, it performs tool and part selection conditioned on the task, the observation, and the extracted object-part candidates.
 
\xhdr{Object and part extraction.} \ours first leverages a VLM to enumerate all objects in the RGB observation $I$, yielding object names $\{o_i\}_{i=1}^{N}$. Then, to enhance the VLM’s understanding of object structure and reduce hallucinations on nonexistent parts, we use SAM3~\cite{SAM3} to segment these objects conditioned on their names, and then extract the corresponding image crops $\{I_i\}_{i=1}^{N}$. For each object name and the corresponding image crop $(o_i, I_i)$, we query the VLM to generate a set of part candidates $P_i = \{p_{i,1}, p_{i,2}, \ldots, p_{i,j_i}\}$, leveraging the VLM's knowledge of object affordances. 

\xhdr{Tool and part selection.} 
Given the RGB image $I$ and the task instruction $l$, we prompt the VLM to select the tool object $o_A$, the target object $o_B$, and task-relevant parts from the extracted object-part candidates $\{(o_i, P_i)\}_{i=1}^{N}$, where $P_i=\{p_{i,1},p_{i,2},\ldots,p_{i,j_i}\}$ denotes the part candidates generated for object $o_i$. We consider two categories of task-relevant parts: (1) a part $p_G$ on the tool, indicating where the robot should grasp $o_A$, and (2) a part $p_A$ on the tool $o_A$ and a part $p_B$ on the target $o_B$, indicating where the two objects should interact to realize the intended function. Concretely, we select $p_G, p_A \in P_A$ and $p_B\in P_B$ in semantic space, and obtain: 
\[\left\{[o_A, (p_G, p_A)], [o_B, p_B]\right\}.\]

The prompts used for object-name extraction, part decomposition, and tool–part selection are provided \hyperref[appendix:prompts]{in Appendix A}.

\subsection{3D Affordance Grounding}\label{subsec:aff_detect}
Now that we have selected the tool object $o_A$, the target object $o_B$, and the task-relevant parts $(p_G, p_A, p_B)$, the next step is to ground each part from the semantic space into a 3D affordance region for manipulation. In practice, these affordance regions may be occluded by other objects in the scene or self-occluded due to the camera viewpoint. Prior work mitigates this by relying on multi-camera setups, which are often costly and slow. Instead, we propose a novel affordance detection approach that produces complete, function-faithful 3D affordance regions from a single-view RGB-D observation. As illustrated in Fig.~\ref{fig:perception}, the proposed approach consists of two stages: (1) 3D reconstruction and registration, and (2) multi-view 2D segmentation and fusion.

\begin{minipage}[ht]{0.48\linewidth}
    \centering
    \captionof{table}{\textbf{Affordance prediction results on AGD20K unseen test split.}}
    \fontsize{8pt}{8pt}\selectfont
    \setlength{\tabcolsep}{5pt}
    \begin{tabular}{lccc}
\toprule
Methods & KLD ($\downarrow$) & SIM ($\uparrow$) & NSS ($\uparrow$) \\
\midrule
Cross-View-AG~\cite{affordance_2d_1}   & 1.787 & 0.285 & 0.829 \\
LOCATE~\cite{locate}          & \textbf{1.405} & 0.372 & 1.157 \\
3DOI~\cite{3DOI}            & 3.565 & 0.227 & 0.657 \\
AffordanceLLM~\cite{affordancellm}   & 1.463 & 0.377 & 1.070 \\
UAD~\cite{uad} & 1.878 & 0.407 & 1.092 \\
\ours & 1.506 & \textbf{0.418} & \textbf{1.334} \\
\bottomrule
\end{tabular}

    \label{tab:exp_agd20k}
    \end{minipage}%
    \hfill
    \begin{minipage}[ht]{0.48\linewidth}
    \centering
    \captionof{table}{\textbf{Affordance prediction results on PIAD unseen test split.}}
    \fontsize{8pt}{8pt}\selectfont
    \setlength{\tabcolsep}{8pt}
    \begin{tabular}{lccc}
\toprule
Methods & aIoU ($\uparrow$) & AUC ($\uparrow$) & MAE ($\downarrow$) \\
\midrule
PFusion~\cite{pfusion} & 5.3 & 61.9 & 0.193 \\
XMF~\cite{xmf} & 5.7 & 62.6 & 0.188\\
IAGNet~\cite{PIAD_bench} & 8.0 & 71.8 & 0.127\\
LASO~\cite{laso}  &8.0 & 69.2 & 0.118  \\
GEAL~\cite{geal}  &8.7 & 72.5 & \textbf{0.102} \\
\ours & \textbf{9.0} & \textbf{74.2} & \textbf{0.102} \\
\bottomrule
\end{tabular}

    \label{tab:exp_PIAD}
\end{minipage}

\xhdr{3D reconstruction and registration.} We first leverage 3D reconstruction for the selected tool object $o_A$ and target object $o_B$ to alleviate occlusions inherent in single-view observations, as shown on the left of Fig.~\ref{fig:perception}. 
For each object $o_i \in \{o_A, o_B\}$, given the single-view RGB-D observation $(I,D)$, we first use SAM3 to segment the object and produce a masked RGB-D crop $(I_i, D_i)$. We then feed the RGB-D crop into SAM3D~\cite{SAM3D} for reconstruction. To preserve the object’s pose and scale in the camera coordinate frame rather than reconstructing an object with arbitrary pose and scale, we convert the depth crop $D_i$ into a point map $X_i$ and use it as a geometric condition during reconstruction. Finally, to further improve geometric accuracy, we align the reconstructed mesh $\mo$ to the original point map $X_i$. Specifically, since $X_i$ is only from a partial observation, we render $\mo$ from the same camera view to produce a partial point map $X_\mo$ and perform ICP-based rigid registration to obtain the transformation $T$ from $X_\mo$ to $X_i$. Applying the transformation $T$ to the reconstructed mesh $\mo$ yields the registered object mesh $\mr$, which is used in the subsequent stage.

\xhdr{Multi-view 2D segmentation and fusion.} 
After acquiring the registered meshes of the tool object and the target object, the next step is to ground the task-relevant parts $p_G, p_A, p_B$ into 3D affordance regions. 
Since grounding parts on open-category 3D objects remains a challenge~\cite{laso}, we aim to leverage the strong 2D grounding capabilities of state-of-the-art vision foundation models.
For each $p \in \{p_G, p_A, p_B\}$, we first render the corresponding mesh $\mr$ from multiple viewpoints and produce a set of 2D images $\{\widehat{I}_k\}_{k=1}^{K}$. This multi-view rendering addresses self-occlusion and ensures the task-relevant part $p$ is visible and can be grounded. In practice, we choose $K=8$.
For each $\widehat{I}_k\in\{\widehat{I}_k\}_{k=1}^{K}$, we ground the description $p$ of the task-relevant part into a 2D segmentation mask $s_k$ using SAM3. 
However, SAM3 may fail to understand some complex text queries~\cite{SAM3}. In this scenario, to improve the reliability of segmentation, we use a VLM~\cite{qwen3vl} as a fallback to translate such descriptions into bounding boxes. Conditioned on these bounding boxes, SAM3 then produces fine-grained segmentations. This design leverages the VLM’s visual-semantic understanding while improving efficiency by minimizing costly VLM queries.

Finally, we aggregate the multi-view 2D segmentations $\{s_k\}_{k=1}^{K}$ into a 3D affordance region. Specifically, for each $s_k\in \{s_k\}_{k=1}^{K}$, we back-project the mask onto the reconstructed mesh $\mr$ using the corresponding camera parameters, yielding a point set on the object surface. We merge the point sets across views by taking the union, yielding an aggregated point set on the object surface. Since the 2D segmentations may be imperfect, the aggregated 3D points could contain noise. We therefore apply DBSCAN~\cite{dbscan} to cluster the points, discard outliers, and retain only the largest cluster as the predicted affordance region.

\section{Experiments}
\label{sec:experiments}

\begin{figure}[ht]
    \centering
    \includegraphics[width=\linewidth]{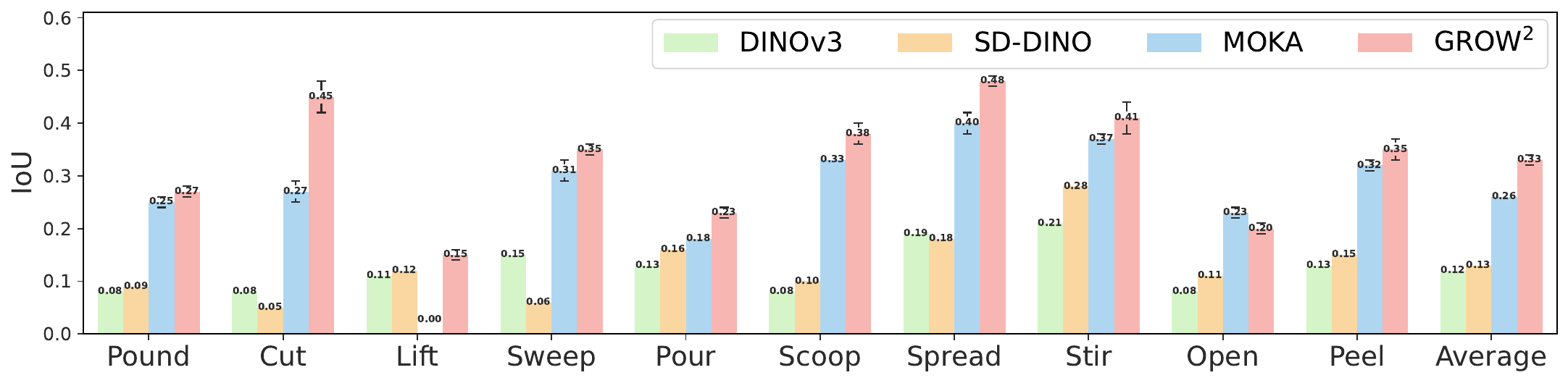}
    \caption{\textbf{Affordance prediction results on \ourbench.} We evaluate alignment between predicted affordances and human annotations with IoU across 10 task types. The results show that \ours aligns better with human experts’ tool selection and affordance annotations.}
     \label{fig:human_study}
\end{figure}

In this section, we aim to answer three questions: (1) How does \ours perform compared to baselines on affordance prediction benchmarks? (2) How effectively do the affordances predicted by \ours support robot tool use? (3) How well does \ours perform in real robot experiments?

\subsection{Affordance Prediction Experiments}
\label{subsec:aff_pred_exp}
\xhdr{\ours achieves superior performance on existing 2D and 3D affordance prediction benchmarks.}
We evaluate \ours on two standard affordance prediction benchmarks: AGD20K~\cite{affordance_2d_1} for 2D images and PIAD~\cite{PIAD_bench} for 3D point clouds. In both benchmarks, the model is required to predict affordance regions conditioned on a task instruction. We evaluate \ours on the unseen test split of each benchmark and compare it against the corresponding baselines. Following the standard evaluation protocols, we report Kullback-Leibler divergence (KLD), similarity metric (SIM), and normalized scanpath saliency (NSS) for AGD20K, and average intersection over union (aIoU), area under the ROC curve (AUC), and mean absolute error (MAE) for PIAD. Details of the baselines and evaluation metrics are provided \hyperref[appendix:affordance_prediction]{in Appendix B}.

Tables~\ref{tab:exp_agd20k} and~\ref{tab:exp_PIAD} show the results on AGD20K and PIAD, respectively.
Despite not being trained on these benchmarks, our method achieves the best overall performance across most metrics, attaining the highest SIM and NSS on AGD20K, as well as the highest aIoU and AUC and the lowest MAE on PIAD. The higher KLD on AGD20K arises because it penalizes mismatches across the full background distribution, while \ours produces fine-grained predictions that may diverge from AGD20K’s more diffuse ground-truth distributions. These baselines directly map task instructions to affordance regions, requiring substantial training data and limiting generalization to open-category objects and novel functions. In contrast, \ours decomposes grounding into two stages, simplifying reasoning and better leveraging existing VLMs and visual foundation models.

\xhdr{\ours can effectively ground affordances in multi-object settings.}
Existing benchmarks, including AGD20K and PIAD, either provide the tool object or involve relatively simple scenes. To evaluate affordance grounding in multi-object settings, we introduce \ourbench, where the robot must first identify which object to use as the tool and then predict its affordance region. Each instance in \ourbench consists of a task instruction and a scene image containing at least five candidate objects, with ground-truth affordance regions annotated by human experts. \ourbench comprises ten task types, with 50 instances each. More details are provided \hyperref[appendix:ourbench]{in Appendix B.3}.
We compare \ours with two types of training-free baselines: (1) one-shot baselines, which select a tool from the scene and transfer the affordance region from a demonstrated tool based on the similarity of pretrained image features, such as DINOv3~\cite{dinov3} and SD-DINO~\cite{sd_dino}; and (2) MOKA~\cite{moka}, a zero-shot baseline that uses a VLM for tool selection and predicts affordances through keypoint-based visual prompting. Implementation details are provided \hyperref[appendix:ourbench_baseline]{in Appendix B.3.1}.

As shown in Fig.~\ref{fig:human_study}, \ours achieves the highest IoU in 9 out of 10 task types. One-shot baselines often fail to select an appropriate tool because pretrained image features lack global semantic understanding. While MOKA uses a VLM for tool selection and affordance prediction, its reliance on visual prompting requires the VLM to reason spatially, making it prone to hallucinations.

\subsection{Simulation Experiments}
\label{subsec:simu_experi}

\begin{table}[t]
\centering
\caption{\textbf{Simulation experiment results.} We report the manipulation success rates (\%) and bold the best results. For VLM-based methods, we report average absolute deviation across three runs. }
\label{tab:exp_mani}
\fontsize{8pt}{8pt}\selectfont
\setlength{\tabcolsep}{11pt}
\begin{tabular}{lcccccc}
\toprule
Method  & Pound      & Cut        & Lift       & Sweep      & Pour       & Average \\
\midrule
ICP~\cite{ICP}    & 35.0       & 40.0       & 5.0        & 55.0       & 20.0       & 31.0    \\
DINOv3~\cite{dinov3}  & 25.0       & 15.0       & 0.0        & 55.0       & 10.0       & 21.0    \\
SD-DINO~\cite{sd_dino} & 15.0       & 30.0       & 0.0        & 70.0       & 65.0       & 36.0    \\
UAD~\cite{uad} (given tool)  & 45.0       & 70.0       & 0.0        & 85.0       & 35.0       & 47.0    \\
GEAL~\cite{geal} (given tool)  & 35.0       & 55.0       & 20.0       & 65.0       & 35.0       & 42.0    \\
MOKA~\cite{moka}  & 36.7\pmm 4.4 & 61.7\pmm 5.6 & 0.0\pmm 0.0  & 50.0\pmm 6.7 & 16.7\pmm 2.2 & 33.0    \\
\ours w/o multi-view & 56.7\pmm 2.2 & 65.0\pmm 3.3 & 60.0\pmm 6.7&71.7\pmm 4.4&60.0\pmm 3.3&	62.7\\
\ours    & \textbf{83.3\pmm 5.6} & \textbf{88.3\pmm 5.6} & \textbf{81.7\pmm 5.6} & \textbf{88.3\pmm 4.4} & \textbf{91.7\pmm 4.4} & \textbf{86.7} \\
\bottomrule
\end{tabular}

\end{table}

\xhdr{Experiment setup.} To evaluate whether \ours can facilitate downstream robot tool use, we construct a simulation benchmark in SAPIEN~3~\cite{sapien} using a Franka Emika Panda arm. The benchmark includes five task types: \textit{Pound}, \textit{Cut}, \textit{Lift}, \textit{Sweep}, and \textit{Pour}, with 20 test cases per task type. Each test case consists of a task instruction and a scene containing one target object and three to five candidate tool objects.

\xhdr{Skill Library.}
To generate executable trajectories from predicted affordances, prior work either learns task-specific policies~\cite{kalm,affordance_rl,part_bc,codediffuser} or uses heuristic constraints for motion planning~\cite{kpam_1,magic,kpam,keto}. We use motion planning as our low-level module, with constraints on geometric features of affordance regions (e.g., 3D bounding boxes and surface normals) for fair comparison. These constraints are instantiated as task-specific key poses. For example, the \textit{pound} task has two key poses: (1) \textit{pre-pound}, where the tool function-region normal points toward the target from above with a prescribed clearance, and (2) \textit{post-pound}, where the tool and target function regions are in contact. Each constraint defines a feasible set of object configurations from which we sample key poses. Definitions for all five task types are provided \hyperref[appendix:low_level]{in Appendix C.1}. To convert these object key poses into executable robot trajectories, we use Contact-GraspNet~\cite{contact_graspnet} to sample candidate grasp poses within the predicted grasp region. Given a selected grasp pose, we derive the corresponding end-effector key poses from the object key poses. We then use a collision-free motion planner (e.g., RRT-Connect~\cite{rrt-connect}) to connect the grasp pose and subsequent end-effector key poses into a complete trajectory. If planning fails, we resample the grasp and object key poses until a valid trajectory is found or a timeout is reached.

\xhdr{Baselines.} We compare \ours with the same baselines as in \ourbench: DINOv3, SD-DINO, and MOKA. We additionally include ICP~\cite{ICP}, a one-shot 3D matching baseline, as well as UAD~\cite{uad} and GEAL~\cite{geal}, the best-performing methods on AGD20K and PIAD, respectively. Since UAD and GEAL are not designed for multi-object scenes, we provide them with the ground-truth tool. We also evaluate \ours w/o multi-view, which ablates the 3D reconstruction and multi-view grounding stage. Additional baseline and implementation details are provided \hyperref[appendix:simulation]{in Appendix C.2}.

\xhdr{Results.} Table~\ref{tab:exp_mani} reports the results. \ours achieves the highest success rate across all five task types, demonstrating strong generalizability across diverse tasks and open-category objects. In particular, \textit{Lift} is the most challenging task for the baselines, as success requires identifying matching affordance regions on both the tool and target objects, with little tolerance for imprecise target affordance localization. In contrast, \ours successfully handles this task through precise semantic reasoning and 3D geometric grounding. Moreover, the performance drop of \ours w/o multi-view highlights the importance of 3D reconstruction and multi-view grounding for resolving occlusions.

\begin{minipage}[b]{0.32\linewidth}
\centering
\includegraphics[width=\linewidth]{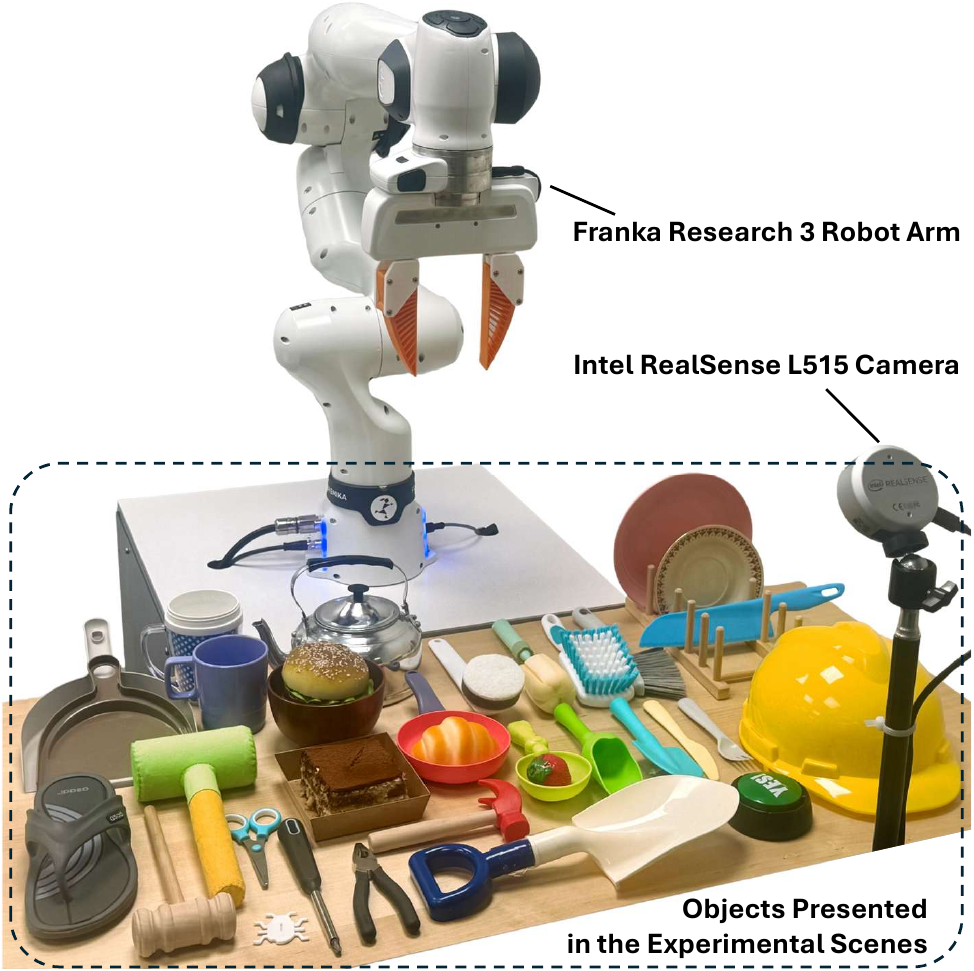}
\captionof{figure}{\textbf{Real-world setup.} }
\label{fig:real_world_setup}
\end{minipage}%
\hfill
\begin{minipage}[b]{0.62\linewidth}
\centering
\includegraphics[width=\linewidth]{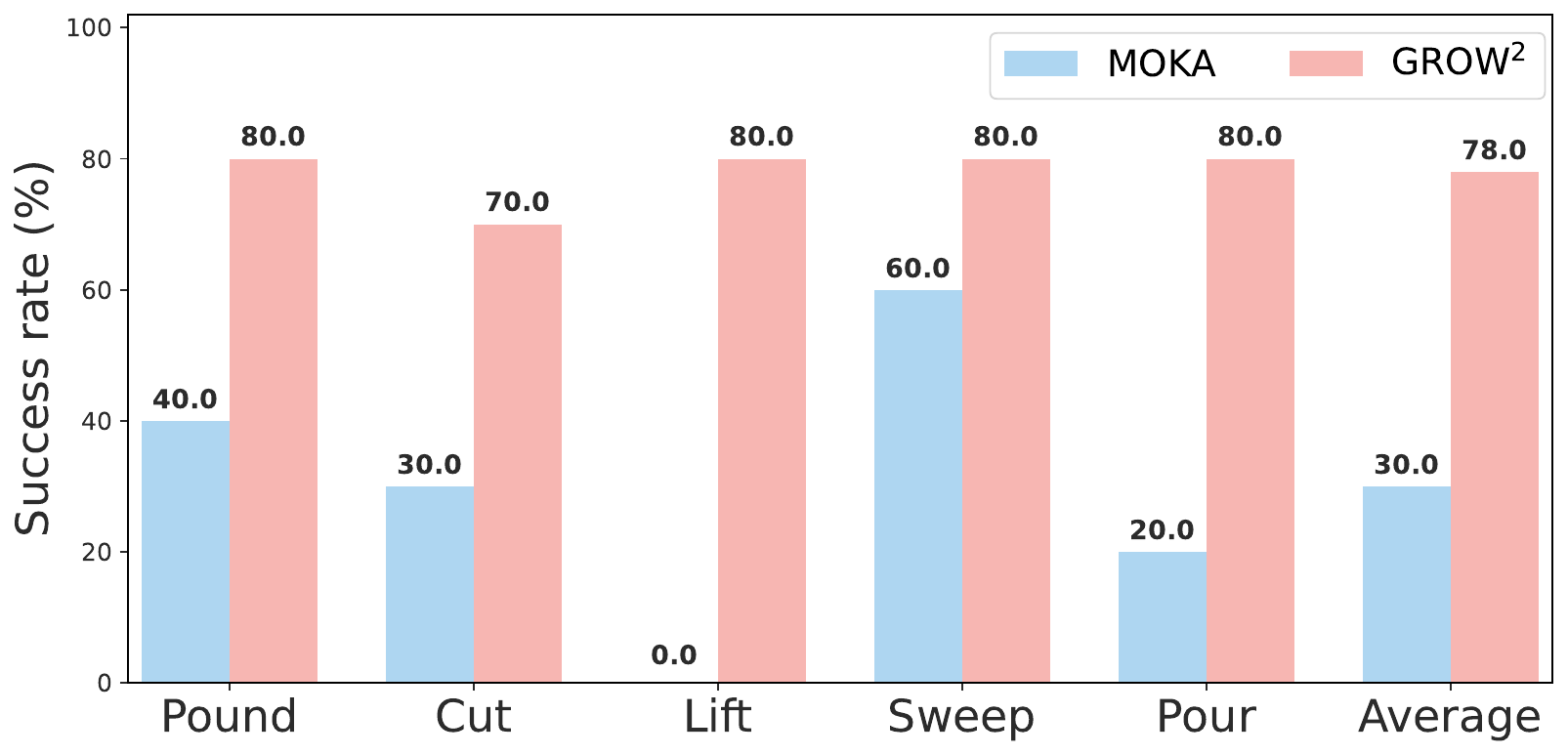}
\captionof{figure}{\textbf{Average success rates (\%) of robot tool use in the real world.} We evaluate on 10 scenes for each task type.}
\label{fig:real_experiment_results}
\end{minipage}

\subsection{Real-World Experiments}
\label{subsec:real_experi}
To validate \ours on a real robotic system, we conduct experiments on five manipulation task types. As shown in Fig.~\ref{fig:real_world_setup}, our setup consists of a Franka Research 3 robot arm with a parallel gripper and an Intel RealSense L515 camera for RGB-D observations. We compare \ours with MOKA~\cite{moka}, which achieves competitive performance in both affordance prediction and simulation experiments. Both methods use the same low-level skills to ensure a fair comparison. For motion planning, we load reconstructed 3D meshes of the tool and target as dynamic bodies and partial point clouds of other objects and the background as static obstacles. For each task type, we report success rates over 10 scenes, each containing a target object and at least five tool candidates.

\begin{figure}[hb]
    \centering
    \includegraphics[width=\linewidth]{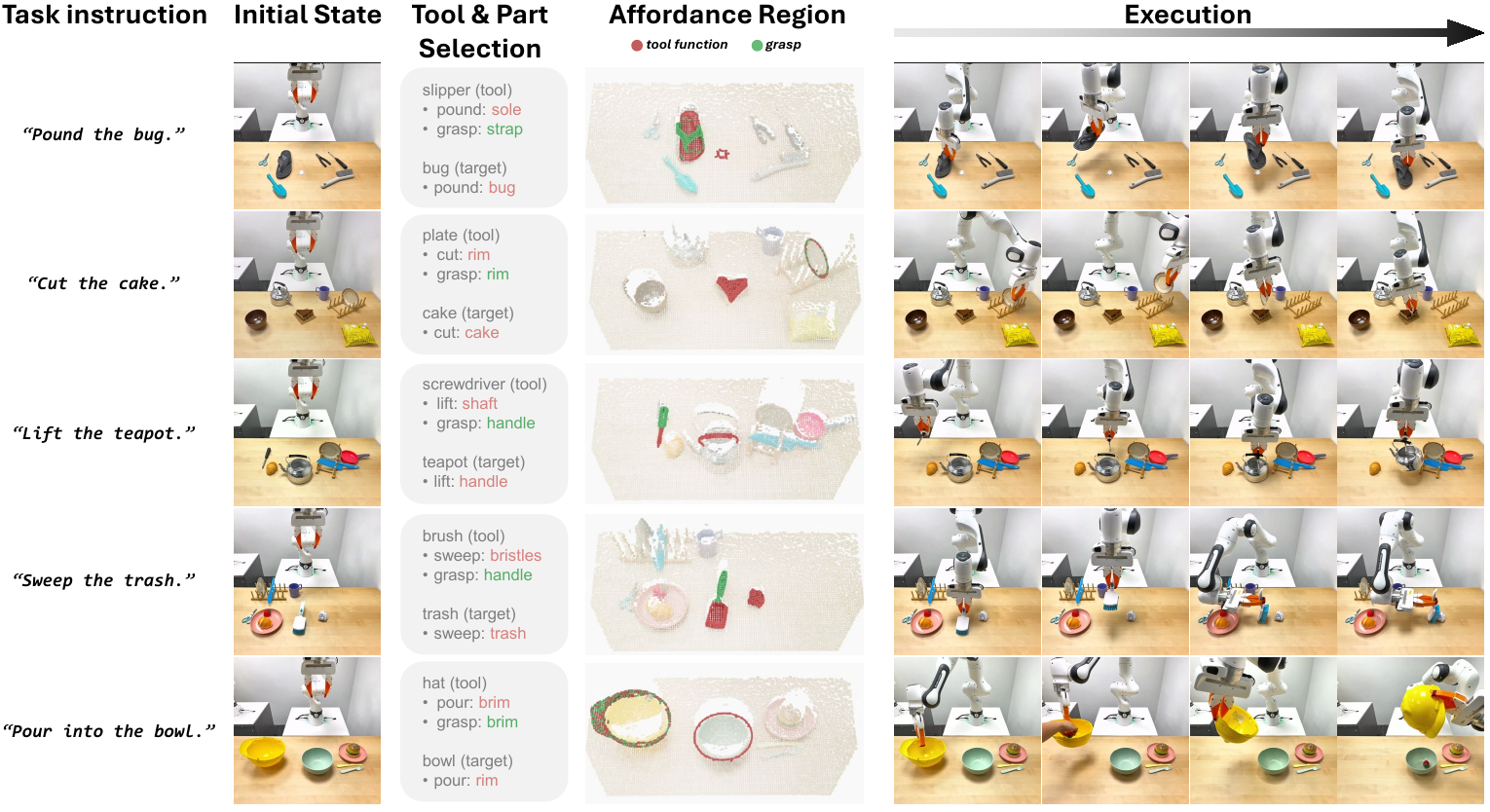}
    \caption{\textbf{Qualitative results of real-world experiments.} }
     \label{fig:real_visual}
\end{figure}

As shown in Fig.~\ref{fig:real_experiment_results}, \ours achieves performance comparable to that in simulation without requiring additional in-domain training data. Fig.~\ref{fig:real_visual} visualizes the predicted affordance regions and the executed trajectories of \ours. Notably, \ours is robust to noisy real-world depth observations, as it computes affordance regions from a reconstructed mesh through accurate multi-view rendering, whereas the baseline is adversely affected by missing or erroneous depth measurements.

\begin{figure}[hb]
    \centering
    \includegraphics[width=\linewidth]{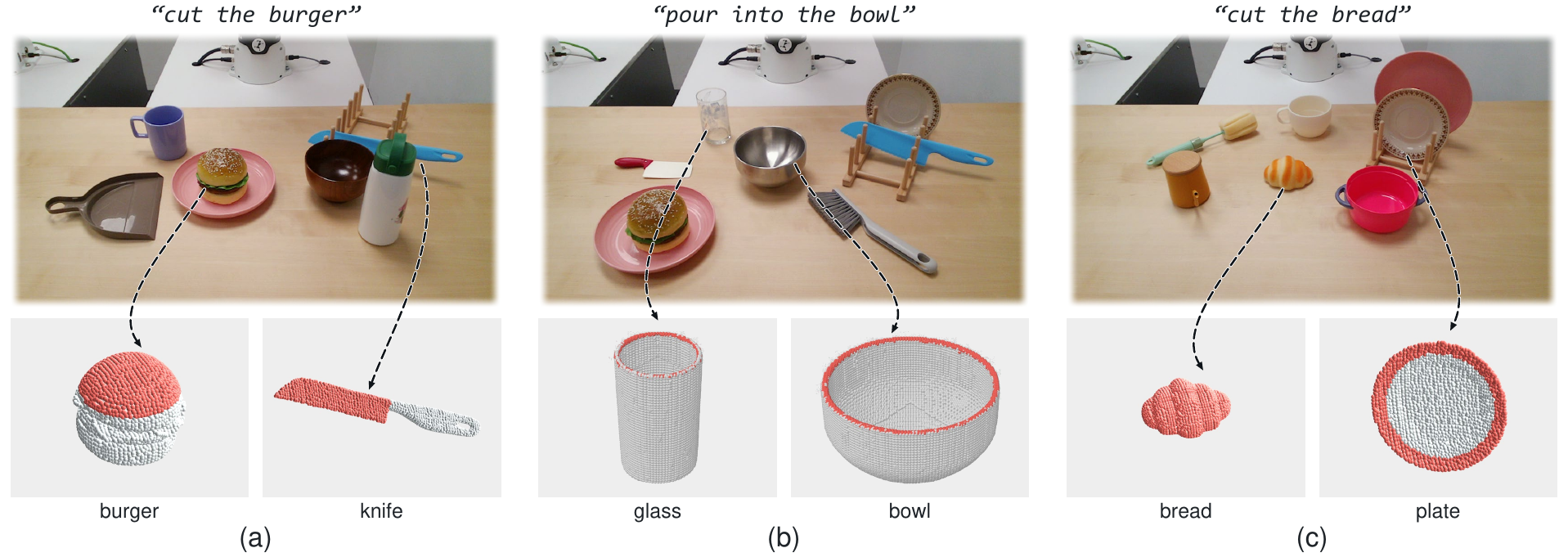}
    \caption{\textbf{Affordance grounding in complex scenes.}
    (a) \textbf{Occlusion:} the burger exhibits self-occlusion, while the knife is occluded by surrounding objects.
    (b) \textbf{Transparent and reflective objects:} affordance grounding on a transparent glass and a reflective bowl.
    (c) \textbf{Multiple tools:} both plates are feasible choices for the task. \textcolor{red}{Red} shows predicted affordance regions.}
    \label{fig:visualization}
\end{figure}

\begin{minipage}[c]{0.48\textwidth}
    \centering
    \renewcommand{\arraystretch}{1.3}
    \setlength{\tabcolsep}{10pt}
    \fontsize{8pt}{8pt}\selectfont

    \begin{tabular}{lcc}
        \toprule
        Model & Time (s) & VRAM (GB) \\
        \midrule
        GPT-5.2 & 2.72 & / \\
        SAM3 & 0.26 & 5.12 \\
        SAM3D & 7.93 & 23.78 \\
        Overall & 16.61 & / \\
        \bottomrule
    \end{tabular}

    \vspace{2mm}

    {\fontsize{7pt}{8pt}\selectfont
    Overall = GPT-5.2$\times$3 +
    SAM3$\times$2 + SAM3D$\times$1}

    \captionof{table}{\textbf{Runtime and memory analysis.}}
    \label{table:running_time}
\end{minipage}
\hfill
\begin{minipage}[c]{0.49\textwidth}
    \centering
    \includegraphics[width=\linewidth]{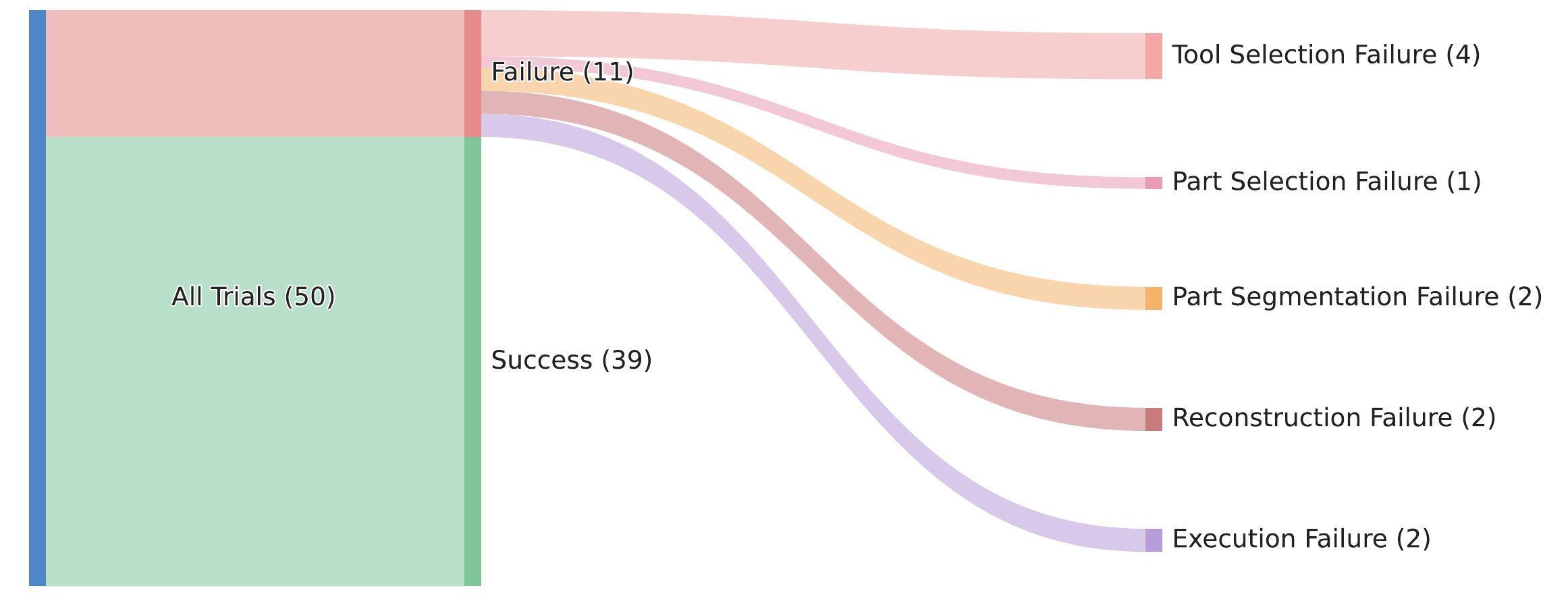}

    \captionof{figure}{\textbf{Failure analysis.}}
    \label{fig:failure_analysis}
\end{minipage}

\ours also demonstrates robustness in complex scenes. In Fig.~\ref{fig:visualization}(a), the burger exhibits self-occlusion, while the knife is partially occluded by surrounding objects. Through 3D reconstruction and multi-view fusion, our method combines complementary observations to recover affordance regions in both cases. Our modular design also facilitates the integration of new models and mechanisms to address additional challenges. For the transparent glass and reflective bowl in Fig.~\ref{fig:visualization}(b), we incorporate Depth Anything 3~\cite{depthanything3} into the perception module to mitigate unreliable depth observations. In Fig.~\ref{fig:visualization}(c), both plates are feasible tools for the task, and the VLM selects a suitable plate based on scene observations and commonsense reasoning. The modular architecture further allows this selection process to be augmented with simulation-based verification and resampling, enabling selection of an alternative tool if the initial candidate fails verification.

\section{Limitations}
\xhdr{Failure cases.}
Fig.~\ref{fig:failure_analysis} presents a breakdown of failure cases across all 50 real-world trials of \ours. Failures arise across four stages: the VLM may select an inappropriate tool or part; SAM3D may produce a mesh with inaccurate scale or geometry; SAM3 may produce an invalid mask for a selected part name; and execution may fail due to improper grasping. Tool and part selection account for the largest share of failures. A promising direction is to wrap the VLM in an agentic system that supports iterative scene inspection, which should mitigate single-shot hallucinations.

\xhdr{Efficiency.} The high latency of VLM queries and vision foundation model inference poses challenges for real-time robotic manipulation. Table~\ref{table:running_time} reports the runtime and memory requirements of \ours, evaluated on an AMD EPYC 9354 CPU and an NVIDIA H100 GPU. To improve computational efficiency, we parallelize object segmentation and 3D reconstruction across tool and target objects, and part segmentation across multiple views. Specifically, each run of \ours involves three GPT-5.2 calls for object-name parsing, part decomposition, and tool-part selection; two SAM3 calls for object and part segmentation, respectively; and one SAM3D call for 3D reconstruction. Future work could distill \ours into a lightweight affordance prediction model~\cite{uad}.

\section{Conclusion}
\label{sec:conclusion}

In this paper, we introduce \ours, a zero-shot method for grounding affordances in robot tool use. By decomposing the problem with a part representation, \ours reformulates open-world affordance grounding as part grounding, enabling existing vision foundation models to address it effectively. \ours can predict complete and functionally faithful 3D affordance regions from a single-view observation. Experiments on affordance prediction benchmarks demonstrate the effectiveness and generalization ability of \ours. Furthermore, tool-use experiments in both simulation and the real world show that the affordance regions predicted by \ours can effectively support downstream robot tool use.

\acknowledgments{
This research is supported in part by the Ministry of Digital Development \& Information, Singapore under the AI Visiting Professorship Program (No. AIVP-2025-008/2025-02181). This research is also supported by the NUS Research Scholarship.}

\bibliography{references}  %

\newpage
\appendix
\section{Prompts}
\label{appendix:prompts}
We use GPT-5.2~\cite{gpt52} (without thinking) as the vision language model (VLM) for object-name parsing, part decomposition, and tool-part selection. Below, we present the system prompts.

\begin{center}
\begin{tcolorbox}[
    colback=gray!5,
    colframe=black!40,
    sharp corners=south,
    title=Object Name Parsing
]
\scriptsize
\raggedright

Identify every visible object in the real-world tabletop workspace and return its normalized category directly. You may receive a full image and overlapping crops of the same scene.

\par\smallskip
No object list or fixed category vocabulary is provided. Recognize objects from visible evidence and use concise, common, lowercase category names suitable for subsequent object detection and segmentation. Use the most specific category supported by the image; when the exact identity is uncertain, use a broader recognizable category instead of guessing. Keep names consistent across views and use the same name for objects of the same category. Return each category once, even if it occurs in multiple crops or multiple instances.

\par\smallskip
Inspect each region systematically, including thin, small, low-contrast and partially occluded objects. Include objects ON, INSIDE, LEANING AGAINST, UNDER, or THROUGH another object or support, not only objects directly touching the tabletop. Look carefully around frames, racks, boxes and holders: a tool resting on a support is a separate object. Follow visible handles, shafts, blades and tips across occlusions; do not split a single tool into its parts or merge it into its support.

\par\smallskip
Use visible support relationships and orientation as recognition context. Check whether an object rests on the table, hangs from a hook or rack, or is held upright by a post or stand. Examine the contact or attachment region and the object's complete shape together. Supports can help identify an object even though the supports themselves must be excluded from the output. Do not infer a support or attachment that is not visible.

\par\smallskip
Before naming an object, check its defining geometry across the full image and crops. Distinguish objects using their complete shape, functional parts and visible context. Do not rely on material, color, a single small detail or an expected task to determine identity. Treat attached handles, blades, bristles, lids and surface decorations as parts of the whole object unless clearly separate.

\par\smallskip
Exclude scene fixtures such as the robot, table, floor, walls, support frames, racks and fixed holders. Include movable containers and other standalone objects when their identity is visually supported. Keep an object's identity separate from its material and location: a wooden-handled tool or a tool on a holder must not be discarded because it is wooden or supported. Do not discard an identifiable object just because it is small or partly occluded. Do not invent fully hidden objects or infer objects from an expected task.

\par\smallskip
Before answering, review each region once more for missed objects and mistaken identities. Combine evidence across all images; the crops repeat objects in the full image. Return only the final normalized category names, without descriptions, fixtures, object parts, explanations or markdown.

\par\smallskip
Return valid JSON in this exact format:

\par\smallskip
\texttt{\{"objects": ["category name", "..."]\}}

\par\smallskip
If no supported objects are visible, return
\texttt{\{"objects": []\}}.

\end{tcolorbox}
\end{center}

\begin{center}
\begin{tcolorbox}[
    colback=gray!5,
    colframe=black!40,
    sharp corners=south,
    title=Part Name Parsing
]
\scriptsize
\raggedright

You are an expert robotic affordance-analysis assistant. You will receive:
\begin{itemize}[nosep,leftmargin=1.2em,itemsep=1pt,topsep=2pt]
    \item an image of an object
    \item a text label for the object: \textless{}object-name\textgreater{}
\end{itemize}

Your task is to list the meaningful physical parts of the object. Output format:

\par\smallskip
\texttt{\{}\\
\hspace*{1em}\texttt{"\textless{}object-name\textgreater{}": [}\\
\hspace*{2em}\texttt{"\textless{}part-1\textgreater{}",}\\
\hspace*{2em}\texttt{"\textless{}part-2\textgreater{}"}\\
\hspace*{1em}\texttt{]}\\
\texttt{\}}

\par\smallskip
Rules:
\begin{itemize}[nosep,leftmargin=1.2em,itemsep=1pt,topsep=2pt]
    \item Use simple, concrete part names (1--2 words).
    \item Parts must be functional object components (e.g., ``handle'', ``rim'', ``tip'', ``blade'', ``shaft'').
    \item All parts must be unique (no duplicates).
    \item Avoid vague and geometry-only terms, such as body, area, wall, edge, surface, base, upper, and interior.
    \item Avoid long or descriptive phrases.
\end{itemize}

If the object does not have meaningful separable parts, return an empty list. After generating the parts, verify that none of them contain any forbidden words. If any do, remove them.

\par\smallskip
Examples:
\begin{itemize}[nosep,leftmargin=1.2em,itemsep=1pt,topsep=2pt]
    \item \texttt{"paintbrush": ["bristles", "handle"]}
    \item \texttt{"ruler": []}
    \item \texttt{"hook": ["shaft", "tip"]}
    \item \texttt{"pan": ["handle", "rim"]}
\end{itemize}

\end{tcolorbox}
\end{center}

\begin{center}
\begin{tcolorbox}[
    colback=gray!5,
    colframe=black!40,
    sharp corners=south,
    title=Tool and Part Selection
]
\scriptsize
\raggedright

You are an expert robot assistant. You will receive:
\begin{itemize}[nosep,leftmargin=1.2em,itemsep=1pt,topsep=2pt]
    \item an image of a tabletop scene
    \item a task instruction describing a manipulation action
    \item an object--part dictionary (each object includes a list of parts)
\end{itemize}

First, reason carefully about the task using:
\begin{itemize}[nosep,leftmargin=1.2em,itemsep=1pt,topsep=2pt]
    \item Geometry---object shapes, orientations, distances, contact regions, and alignment between tool and target
    \item Semantics---object categories, functions, and affordances relevant to the action
    \item Physics---stability, friction, force transfer, leverage, support, and motion constraints
\end{itemize}

Use this reasoning to decide:
\begin{itemize}[nosep,leftmargin=1.2em,itemsep=1pt,topsep=2pt]
    \item target object: identify the object explicitly mentioned or implied as being acted upon in the task instruction.
    \item tool object: select a different object that can perform the required action on the target object, based on its shape and affordances.
    \item \texttt{target\_function\_part}: the part of the target object that will be acted on.
    \item \texttt{tool\_grasp\_part}: the part of the tool object that the robot should grasp.
    \item \texttt{tool\_function\_part}: the part of the tool that performs the action on the target.
\end{itemize}

For some objects that cannot be decomposed, the grasp part and action part can be the object itself, like a ruler. The target and tool objects must be different objects. The tool function part could be the same as the tool grasp part.

\par\smallskip
Your response must contain two parts: (i) explain your reasoning and selection based on geometry, semantics, and physics. You should make use of object parts for reasoning. For example, a container's rim or brim can serve as the function part for pouring because liquid flows over that edge. (ii) A JSON block containing only the final selection results, placed strictly between a pair of \texttt{***} markers.

\par\smallskip
Return your answer only in the following format:

\par\smallskip
\texttt{reason: \textless{}reasoning of tool and part selection\textgreater{}}\par
\texttt{***}\par
\texttt{\{}\par
\hspace*{1em}\texttt{"target\_object": "\textless{}target-object-name\textgreater{}",}\par
\hspace*{1em}\texttt{"target\_object\_info": \{}\par
\hspace*{2em}\texttt{"target\_function\_part": "\textless{}target-part\textgreater{}"}\par
\hspace*{1em}\texttt{\},}\par
\hspace*{1em}\texttt{"tool\_object": "\textless{}tool-object-name\textgreater{}",}\par
\hspace*{1em}\texttt{"tool\_object\_info": \{}\par
\hspace*{2em}\texttt{"tool\_grasp\_part": "\textless{}tool-part-to-grasp\textgreater{}",}\par
\hspace*{2em}\texttt{"tool\_function\_part": "\textless{}tool-part-used-to-act\textgreater{}"}\par
\hspace*{1em}\texttt{\}}\par
\texttt{\}}\par
\texttt{***}

\end{tcolorbox}
\end{center}

\section{Details of Affordance Prediction Experiments}
\label{appendix:affordance_prediction}
\subsection{AGD20K}
AGD20K~\cite{affordance_2d_1} is a 2D affordance grounding benchmark, consisting of 20,061 exocentric images and 6,060 egocentric images from 36 affordance categories. We follow the evaluation procedure defined in AGD20K~\cite{affordance_2d_1} and compare \ours with baselines on the unseen test split using the same metrics as in previous work.

\subsubsection{Baselines}
\xhdr{Cross-View-AG}~\cite{affordance_2d_1} learns affordance cues from diverse exocentric interaction images and transfers them to the egocentric view. It further improves affordance localization by preserving correlations among affordance regions.

\xhdr{LOCATE}~\cite{locate} learns affordance grounding from weak supervision by identifying the object part involved in exocentric human-object interactions and transferring this part-level cue to egocentric object images.

\xhdr{3DOI}~\cite{3DOI} predicts affordance regions from a single RGB image by identifying interactive objects and localizing where human-object interactions are likely to occur.

\xhdr{AffordanceLLM}~\cite{affordancellm} uses LLaVA as a vision-language backbone to understand the affordance, then predicts the affordance region through an added mask decoder and mask token.

\xhdr{UAD}~\cite{uad} extracts knowledge from foundation models and distills it into a lightweight affordance prediction model without manual annotation.

\subsubsection{Metrics}

\textbf{K}ullback-\textbf{L}eibler \textbf{D}ivergence (\textbf{KLD}) measures the distributional difference between the predicted affordance map ($M$) and the ground truth ($M'$), which is 
\begin{equation}
   \mathrm{KLD}\left ( M,M' \right )=\sum_{i}M'_{i}\log\left ( \epsilon + \frac{M'_{i}}{\epsilon+M_{i}} \right ), \label{eq:no20}
\end{equation}
where $\epsilon$ is a small positive constant added for numerical stability.

\textbf{Si}milarity \textbf{M}etric (\textbf{SIM}) is also called histogram intersection, which measures the intersection between the predicted affordance map ($M$) and the ground truth ($M'$). The score ranges from 0 to 1. It is given by
\begin{equation}
   \mathrm{SIM}\left ( M, M' \right )=\sum_{i}\min\left ( M_{i},M'_{i}\right ),\\
\end{equation}
where  $\sum_{i}M_{i}=\sum_{i}M'_{i}=1$.

\textbf{N}ormalized \textbf{S}canpath \textbf{S}aliency (\textbf{NSS}) measures the correspondence between the prediction map ($M$) and the ground truth ($M'$). It is given by
\begin{equation}
   \mathrm{NSS}\left ( M,M' \right )=\frac{1}{N}\sum_{i}\hat{M}_{i}\times M'_{i}, \label{eq:no22}
\end{equation}
where $N=\sum_{i}M'_{i}$, $\hat{M}=\frac{M-\mu\left ( M \right )}{\sigma\left ( M \right )}$. $\mu\left ( M \right )$ and $\sigma\left ( M \right )$ are the mean and standard deviation, respectively.

\subsubsection{Qualitative Results}
\begin{figure}[th]
  \centering
  \includegraphics[width=\linewidth]{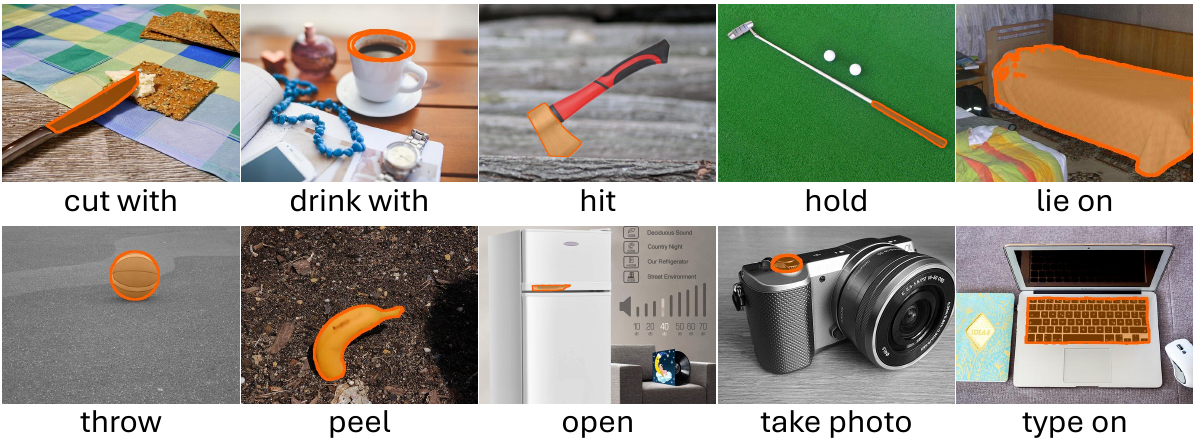}
  \caption{\textbf{Representative examples from the AGD20K benchmark.} We visualize the affordances predicted by \ours using the orange regions in the images.}
  \label{fig:agd20k}
\end{figure}

In Fig.~\ref{fig:agd20k}, we show representative qualitative results on the unseen split of AGD20K. Across diverse affordance categories such as cutting, drinking, opening, and typing, \ours consistently highlights task-relevant image regions, indicating its ability to perform affordance grounding on open-category objects.

\subsection{PIAD}
PIAD~\cite{PIAD_bench} is a 3D affordance grounding benchmark, consisting of 7,012 point clouds across 23 object classes and 17 affordance categories. We follow the evaluation protocol defined in PIAD and compare \ours with baselines on the unseen test split using the same metrics as in previous work. Since the point clouds in PIAD do not contain color information, it is difficult to leverage semantic cues for affordance grounding. In realistic settings, however, RGB images are typically available when capturing point clouds. To evaluate \ours on this benchmark, we make the following modification. After multi-view rendering, we obtain a set of depth images. We then apply ControlNet~\cite{controlnet_depth} to generate RGB images conditioned on these depth images, and perform part segmentation on the generated RGB images. The rest of the pipeline remains unchanged.

\subsubsection{Baselines}

\xhdr{PFusion}~\cite{pfusion} is a multimodal fusion baseline that combines global image features, global point-cloud features, and point-wise 3D features for dense 3D prediction.

\xhdr{XMF}~\cite{xmf} performs cross-modal feature fusion with modality-specific feature extraction and attention-based interaction between 2D and 3D representations.

\xhdr{IAGNet}~\cite{PIAD_bench} is the original PIAD baseline, which grounds 3D affordance regions by aligning interaction-aware image features with point-cloud features.

\xhdr{LASO}~\cite{laso} uses an adaptive fusion module to integrate language and point-cloud features for language-conditioned 3D affordance region prediction.

\xhdr{GEAL}~\cite{geal} predicts 3D affordance regions from semantic cues by bridging sparse 3D point clouds with 2D representations via Gaussian splatting.

\subsubsection{Metrics}

\textbf{A}rea \textbf{U}nder the ROC \textbf{C}urve (\textbf{AUC}) measures how well the predicted affordance map separates affordance and non-affordance regions over different thresholds.

\textbf{A}verage \textbf{I}ntersection over \textbf{U}nion (\textbf{aIoU}) measures the average overlap between the predicted affordance region and the ground-truth region:
\begin{equation}
    \mathrm{IoU} = \frac{\mathrm{TP}}{\mathrm{TP}+\mathrm{FP}+\mathrm{FN}},
\end{equation}
where $\mathrm{TP}$, $\mathrm{FP}$, and $\mathrm{FN}$ denote true positive, false positive, and false negative counts, respectively.

\textbf{M}ean \textbf{A}bsolute \textbf{E}rror (\textbf{MAE}) measures the average absolute difference between the predicted affordance map and the ground truth:
\begin{equation}
\mathrm{MAE}=\frac{1}{n} \sum_{i=1}^n\left|e_i\right|,
\end{equation}
where $e_i$ denotes the prediction error for the $i$-th element.

\subsubsection{Qualitative Results}
\begin{figure}[h]
  \centering
  \includegraphics[width=\linewidth]{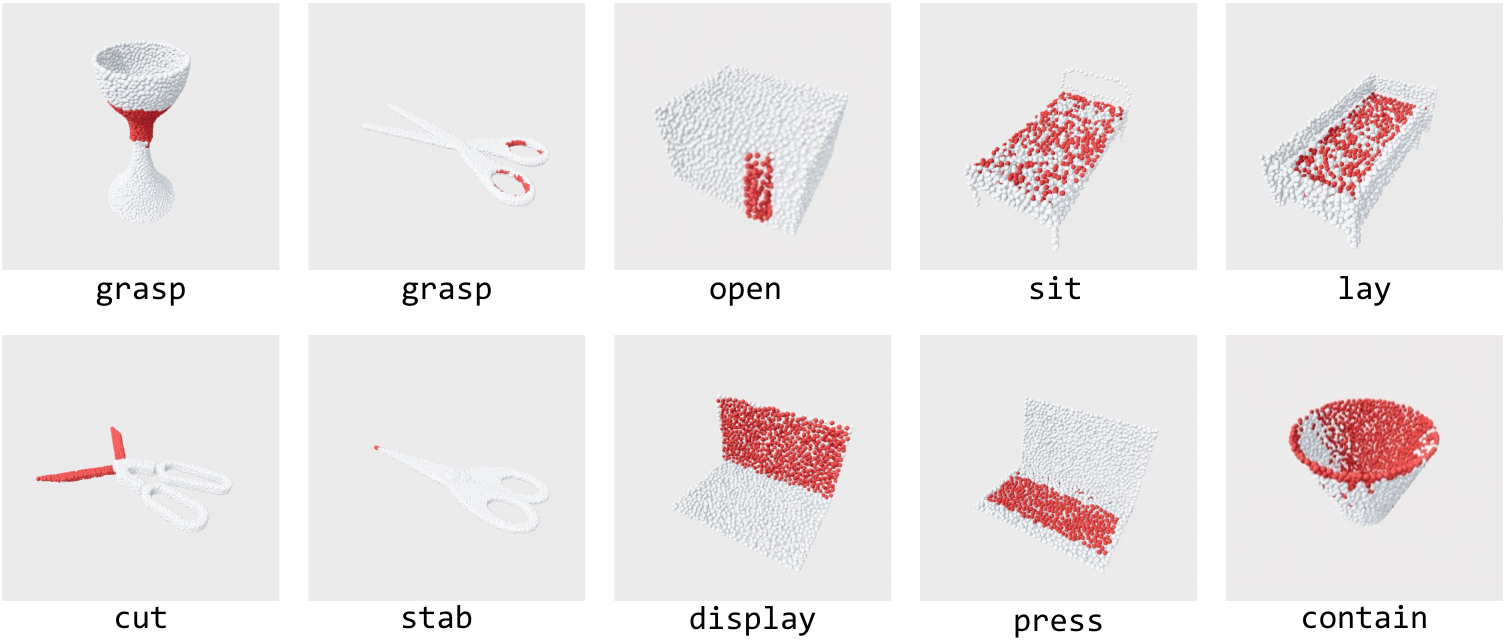}
  \caption{\textbf{Representative examples from the PIAD benchmark.} We visualize the affordances predicted by \ours with the red regions in the point cloud.}
  \label{fig:piad}
\end{figure}

Fig.~\ref{fig:piad} presents qualitative examples on the unseen split of PIAD. The results show that \ours can localize meaningful 3D affordance regions across different object classes and interaction types, demonstrating the effectiveness of \ours on 3D affordance grounding.

\subsection{\ourbench}
\label{appendix:ourbench}
Existing benchmarks, including AGD20K and PIAD, either provide the tool object or involve relatively simple scenes. To evaluate affordance grounding in multi-object settings, we introduce \ourbench, where the robot must first identify which object to use as the tool and then predict its affordance region. The construction procedure for \ourbench is as follows.

First, we prompt Gemini 3 Pro~\cite{gemini3pro} to generate a comprehensive inventory of household objects, resulting in 82 object categories. Next, for each of the 10 task types, we define a set of plausible target objects containing 4–8 candidates. We then ask Gemini 3 Pro to propose a list of possible tool objects for each task type. This list is subsequently verified and filtered by a human annotator, yielding 10–20 tool candidates per task type. For each test scene, we randomly sample (i) one target object from the task-specific target set, (ii) one tool object from the corresponding tool-candidate set, and (iii) four distractor objects from the household object inventory. We then use Nano Banana Pro~\cite{google2026nanobanana} to generate a tabletop image containing these six objects, ensuring that the target object referenced in the instruction is placed near the image center. For each of the 10 task types, we generate 50 scenes, resulting in 500 test scenes in total.

\begin{figure}[ht]
  \centering
  \includegraphics[width=\linewidth]{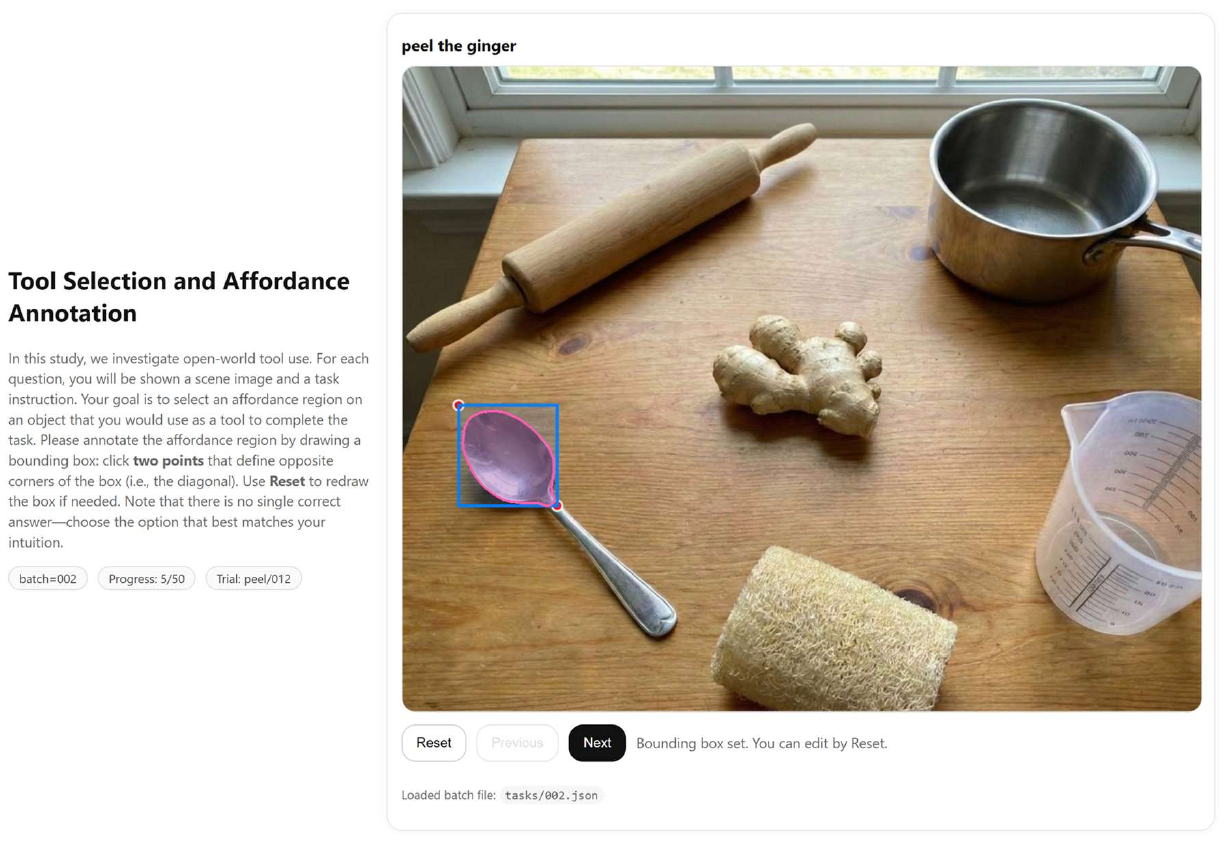}
  \caption{\textbf{Web-based annotation interface for the construction of \ourbench.}}
  \label{fig:web}
\end{figure}

\begin{figure}[t]
  \centering
  \includegraphics[width=\linewidth]{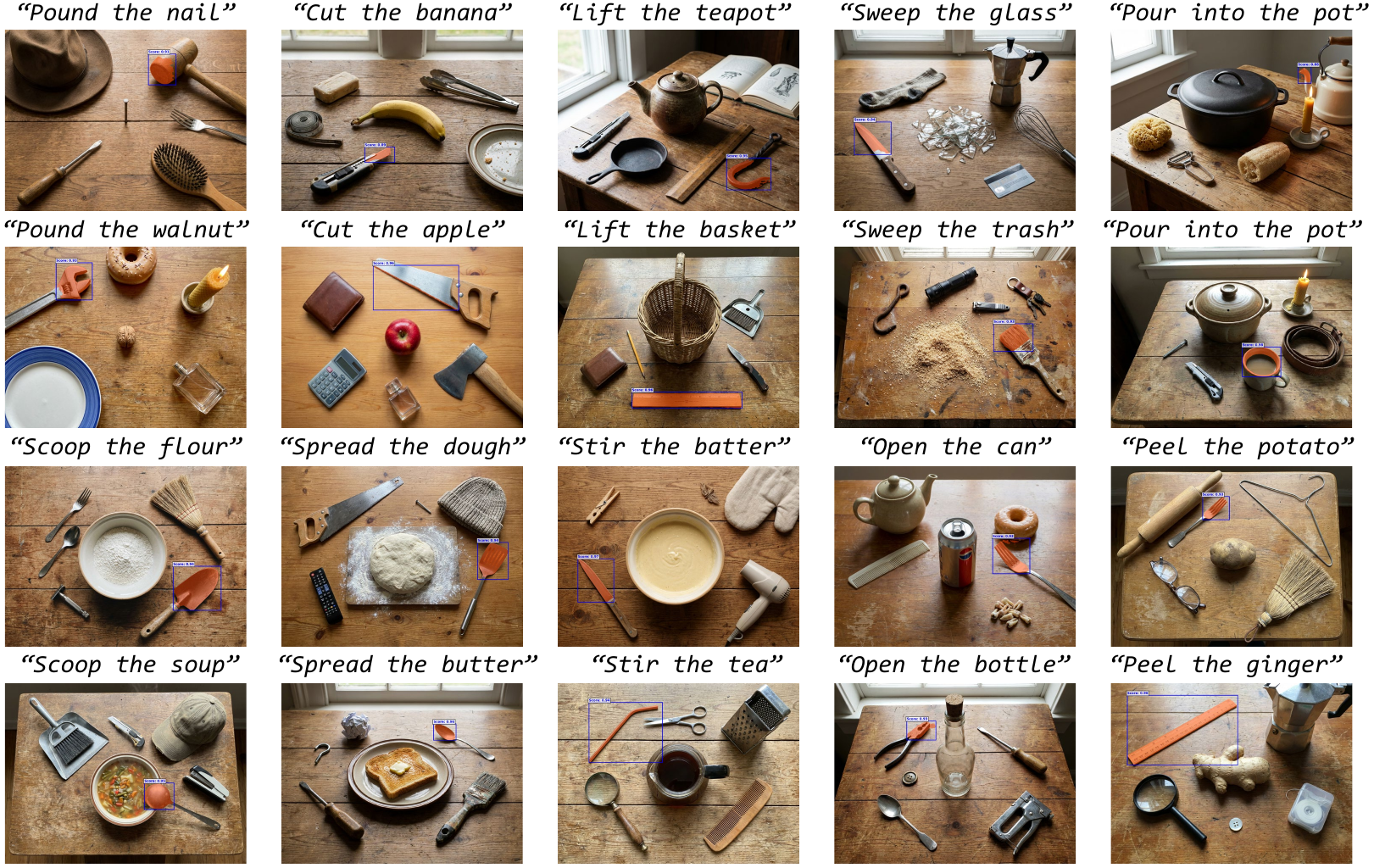}
  \caption{\textbf{Representative examples from \ourbench.} We visualize ground-truth affordance regions annotated by human experts with the blue bounding boxes and the orange regions.}
  \label{fig:ourbench}
\end{figure}

We additionally build a web-based annotation interface to collect human selections of the tool object and the corresponding affordance region, as shown in Fig.~\ref{fig:web}. We first present several examples to familiarize annotators with the procedure. Given an image and its paired task instruction, a robotics expert selects the appropriate tool object and draws a bounding box around the tool’s affordance region. We convert each bounding box into a 2D segmentation mask using SAM 3, which serves as the annotated affordance region. To reduce individual bias, we recruit 10 robotics experts and aggregate their annotations. Fig.~\ref{fig:ourbench} shows representative examples from the dataset.

\subsubsection{Baselines}
\label{appendix:ourbench_baseline}
\begin{figure}[t]
  \centering
  \includegraphics[width=\linewidth]{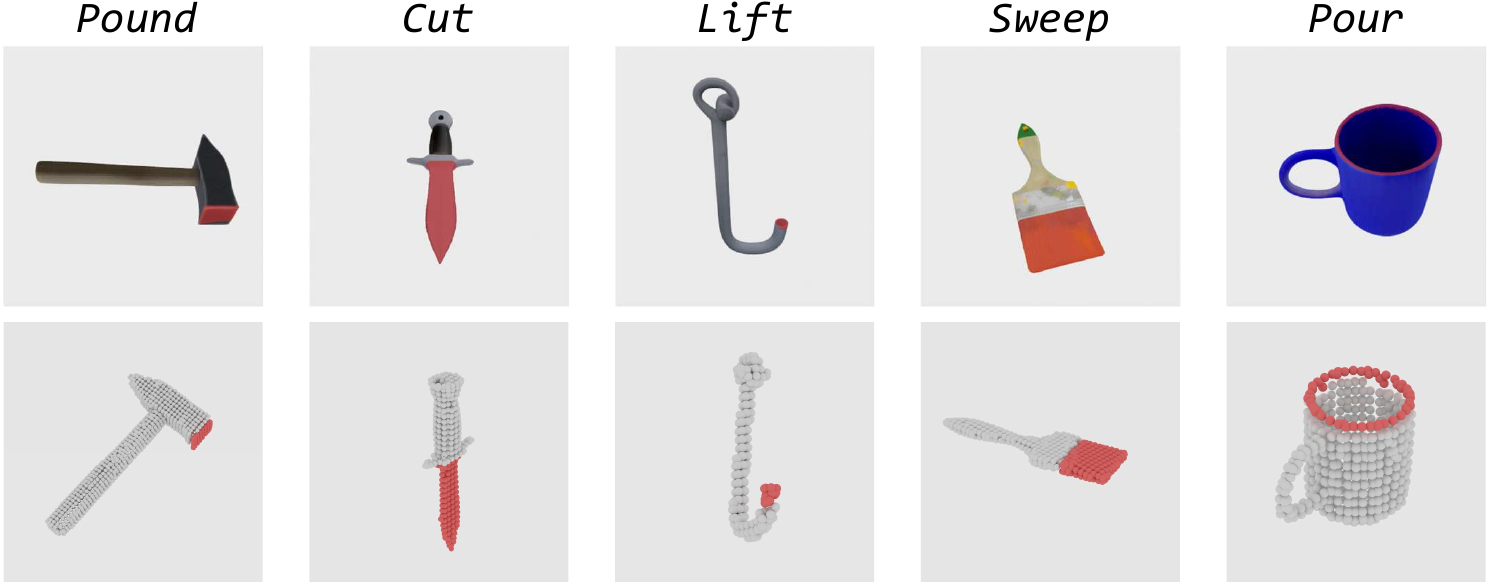}
  \caption{\textbf{Demonstrations for one-shot baseline methods.} For each task category, we provide one demonstration. The affordance region (shown in red) is annotated on both 2D images and 3D point clouds.}
  \label{fig:demo}
\end{figure}

 We implement several training-free methods as baselines for \ourbench and categorize them into one-shot and zero-shot settings. The one-shot baselines rely on a demonstration and feature matching to select the tool and localize the affordance region, whereas the zero-shot baselines leverage the semantic and spatial understanding capabilities of the VLM. For the one-shot baselines, we provide one demonstration per task category. Each demonstration consists of observations of a conventional tool in both a 2D image and a 3D point cloud, along with an annotated affordance region on the tool (Fig.~\ref{fig:demo}). For example, in the pounding task, the conventional tool is a hammer, and we annotate the affordance region on both the 2D image and the 3D point cloud. Details of the one-shot baselines are as follows.

\begin{figure}[th]
  \centering
  \includegraphics[width=\linewidth]{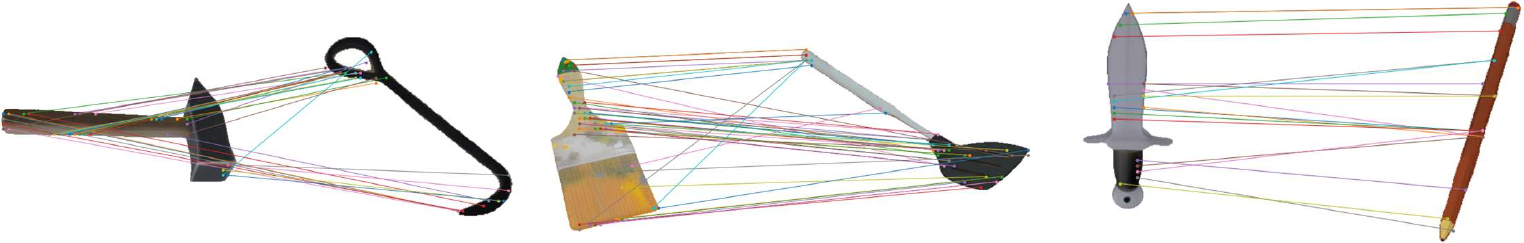}
  \caption{\textbf{Best-Buddies Similarity (BBS)~\cite{bbs} matching with SD-DINO~\cite{sd_dino} image features.}}
  \label{fig:corr}
\end{figure}

\xhdr{SD-DINO}. We first use a VLM to enumerate the objects in the scene and obtain their names. We then use SAM3~\cite{SAM3} to segment each object conditioned on its name and extract the corresponding image crops. Next, we match each crop to the demonstrated tool image and select the crop with the highest similarity score as the tool. To extract matching features, we use SD-DINO~\cite{sd_dino}, which combines Stable Diffusion~\cite{stable_diffusion} and DINOv2~\cite{dinov2} representations to capture both global context and local details. To further improve matching robustness, we compute the Best-Buddies Similarity (BBS)~\cite{bbs} using SD-DINO features. BBS measures structural similarity by identifying pairs of local patches—one from the image crop and one from the demonstration image—that are mutual nearest neighbors in feature space (Fig.~\ref{fig:corr}). These “best-buddy” pairs capture strong local correspondences. We compute the similarity for each matched pair and use the sum of pairwise similarities as the overall matching score. The crop (and the object) with the highest score is selected. To localize the affordance region, we transfer the demonstration affordance to the selected image crop using the best-buddy correspondences: we take the matched patch pairs whose demonstration-side patches fall inside the annotated affordance region, and map their corresponding patches onto the selected image crop to obtain the affordance region on the selected tool.

\xhdr{DINOv3}. The DINOv3 baseline is implemented in the same way as SD-DINO, except that it uses a different matching feature. DINOv3~\cite{dinov3} has been pretrained on internet-scale data and can provide rich, transferable visual features.

The zero-shot baseline is \textbf{MOKA}~\cite{moka}, which uses a VLM to select the tool and target object, and to propose affordance keypoints via visual prompting. To improve MOKA’s tool-selection performance, we first use a VLM to enumerate the objects in the scene and obtain their names. We then query the VLM for tool selection using the object names together with the RGB image. After selecting the tool and target objects, MOKA applies farthest point sampling to sample candidate keypoints along the object boundary, and queries the VLM to select the grasp, function, and target keypoints. Finally, to obtain the affordance region, we feed the selected keypoints to SAM3 and use the resulting segmentation mask as the affordance region.

\subsubsection{Metrics}
We use \textbf{I}ntersection over \textbf{U}nion (\textbf{IoU}) to measure the average overlap between the predicted affordance region and the human-annotated affordance region:
\begin{equation}
    \mathrm{IoU} = \frac{\mathrm{TP}}{\mathrm{TP}+\mathrm{FP}+\mathrm{FN}},
\end{equation}
where $\mathrm{TP}$, $\mathrm{FP}$, and $\mathrm{FN}$ denote true positive, false positive, and false negative counts, respectively.

\subsubsection{Qualitative Results}

\begin{figure}[h]
  \centering
  \includegraphics[width=\linewidth]{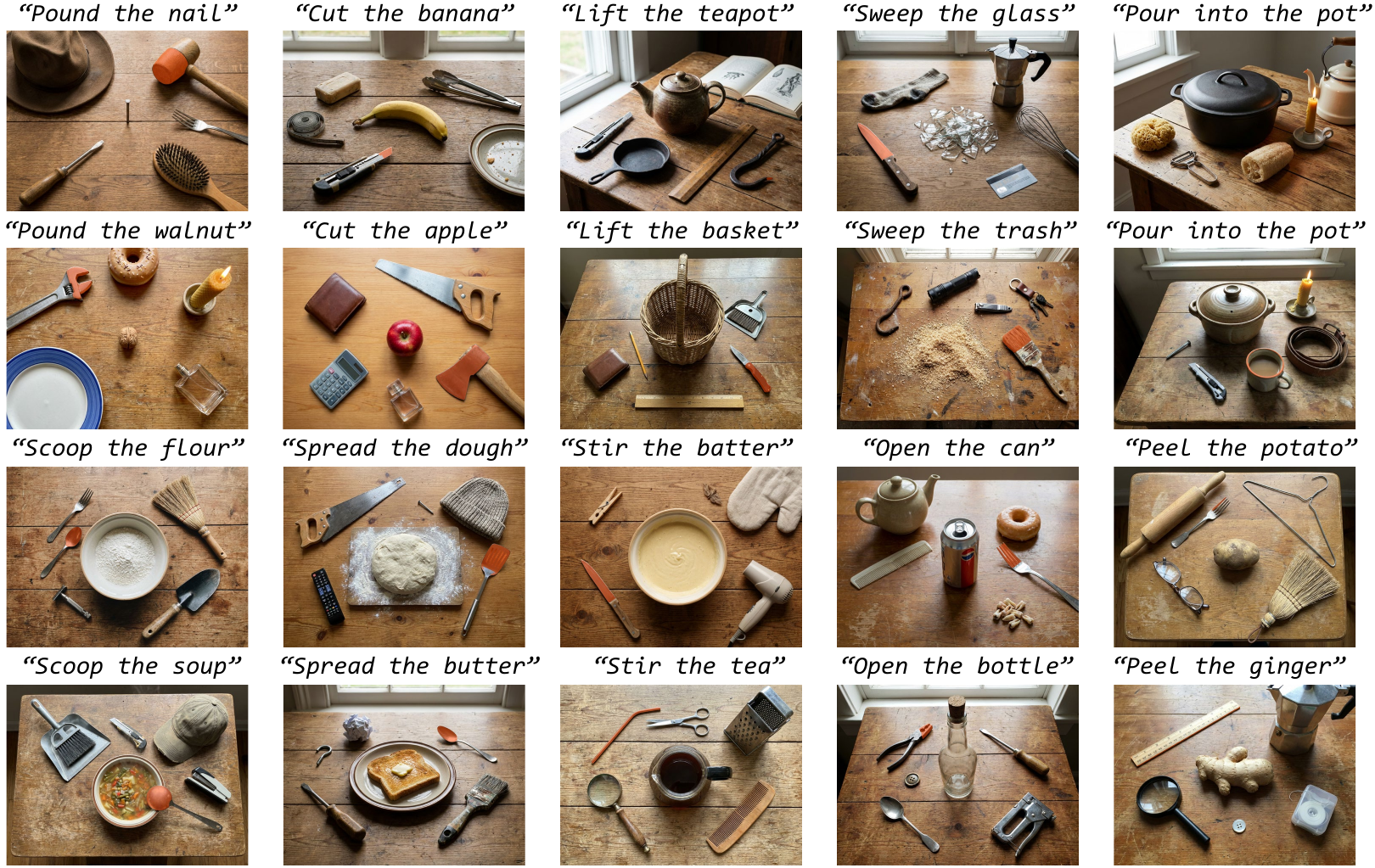}
  \caption{\textbf{Affordance prediction results on \ourbench.} We visualize the affordances predicted by \ours with the orange regions in the images.}
  \label{fig:result_ourbench}
\end{figure}

Fig.~\ref{fig:result_ourbench} visualizes qualitative results on \ourbench. In contrast to AGD20K and PIAD, each scene contains multiple candidate objects, requiring the model to first select an appropriate tool and then localize its functional region. The examples show that \ours can identify appropriate tools and predict consistent affordance regions across a wide range of tool-use tasks.

\section{Details of Robot Tool Use Experiments}
\subsection{Key Poses for Low-Level Skills}
\label{appendix:low_level}
\begin{figure}[h]
  \centering
  \includegraphics[width=\linewidth]{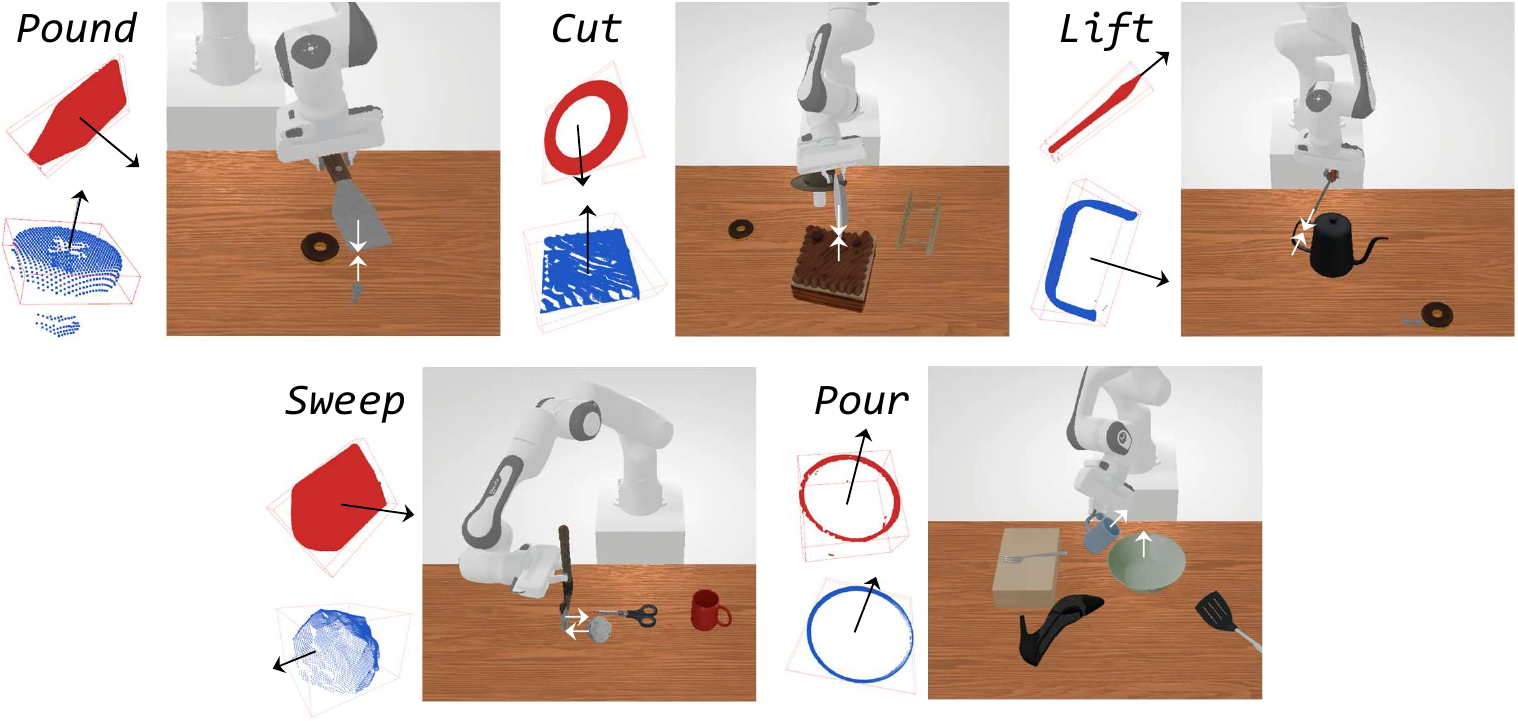}
  \caption{\textbf{Representative key poses and region-based constraints for the low-level skills.}}
  \label{fig:motion_planning}
\end{figure}

We define task-specific key poses using geometric constraints over predicted affordance regions. Here, we provide the detailed key-pose definitions and constraints for all five task types. Fig.~\ref{fig:motion_planning} illustrates a representative key pose for each task, with the tool and target affordance regions shown in the upper-left and lower-left panels, respectively, and the corresponding key pose on the right. The key poses and their associated constraints are defined as follows:

\begin{itemize}[nosep,leftmargin=1.2em,itemsep=1pt,topsep=2pt]
\item \textbf{Pound}:
The pounding task consists of two key poses.
(1) In \textit{pre-pound}, the normal vector of the tool function region is aligned with and opposite to the normal vector of the target function region, while the two regions are separated by a fixed offset distance.
(2) In \textit{post-pound}, the two normals remain aligned and opposite, and the two function regions are brought into contact.

\item \textbf{Cut}: 
We first sample a \emph{cutting vector} from the tool function region. Specifically, the vector originates from the center of the oriented bounding box (OBB) of the tool function region and is perpendicular to the shortest OBB edge.
The cutting task consists of two key poses.
(1) In \textit{pre-cut}, the cutting vector is aligned with and opposite to the normal vector of the target function region, with a fixed offset distance between the two regions.
(2) In \textit{post-cut}, the two vectors remain aligned and opposite, and the two function regions intersect.

\item \textbf{Lift}: 
We first define an \emph{insertion vector} for the tool and a \emph{socket vector} for the target. The insertion vector is aligned with the longest edge of the OBB of the tool function region, and its origin is placed at the endpoint of the OBB along that edge. The socket vector represents the insertion axis of the target; it is aligned with the shortest edge of the OBB of the target function region and originates from the OBB center.
The lifting task consists of three key poses.
(1) In \textit{pre-insert}, the insertion vector is aligned with the socket vector, and the two vector origins are separated by a fixed offset distance.
(2) In \textit{post-insert}, the tool moves a fixed distance from the pre-insert pose along the insertion vector.
(3) In \textit{post-lift}, the tool moves a fixed distance from the post-insert pose along the positive $z$ direction.

\item \textbf{Sweep}: 
We first sample a surface normal from the target function region that is consistent with the specified sweep direction.
The sweeping task consists of two key poses.
(1) In \textit{pre-sweep}, the normal vector of the tool function region is aligned with and opposite to the sampled vector on the target function region, with a fixed offset distance between the two regions.
(2) In \textit{post-sweep}, the tool moves a fixed distance from the pre-sweep pose along the sweep direction.

\item \textbf{Pour}: 
We first extract two task-relevant regions: a \emph{receiving region} on the target and a \emph{pour region} on the tool. The receiving region is represented as a rim circle extracted from the target function region, with its center at the rim center and its radius estimated from the target bowl AABB in the $xy$ plane. The pour region is extracted as the spout or end-cap of the tool function region. We define the principal \emph{pour direction} as the shortest OBB axis of the pour region, oriented to point outward. We then sample a \emph{pour point} on the pour region, biased away from the grasp point to avoid pouring toward the gripper.
The pouring task consists of two key poses.
(1) In \textit{pre-pour}, the sampled pour point is placed above the target rim with a fixed height offset and laterally positioned near the rim along a fixed approach direction. The tool is oriented such that the pour direction is tilted toward the target by a small angle, e.g., $30^\circ$.
(2) In \textit{post-pour}, the pour point remains at the same rim-relative position, while the tool rotates to a larger tilt angle, e.g., $90^\circ$, to direct the flow into the target.

\end{itemize}

\subsection{Simulation Experiments}
\label{appendix:simulation}

\begin{figure}[t]
  \centering
  \includegraphics[width=\linewidth]{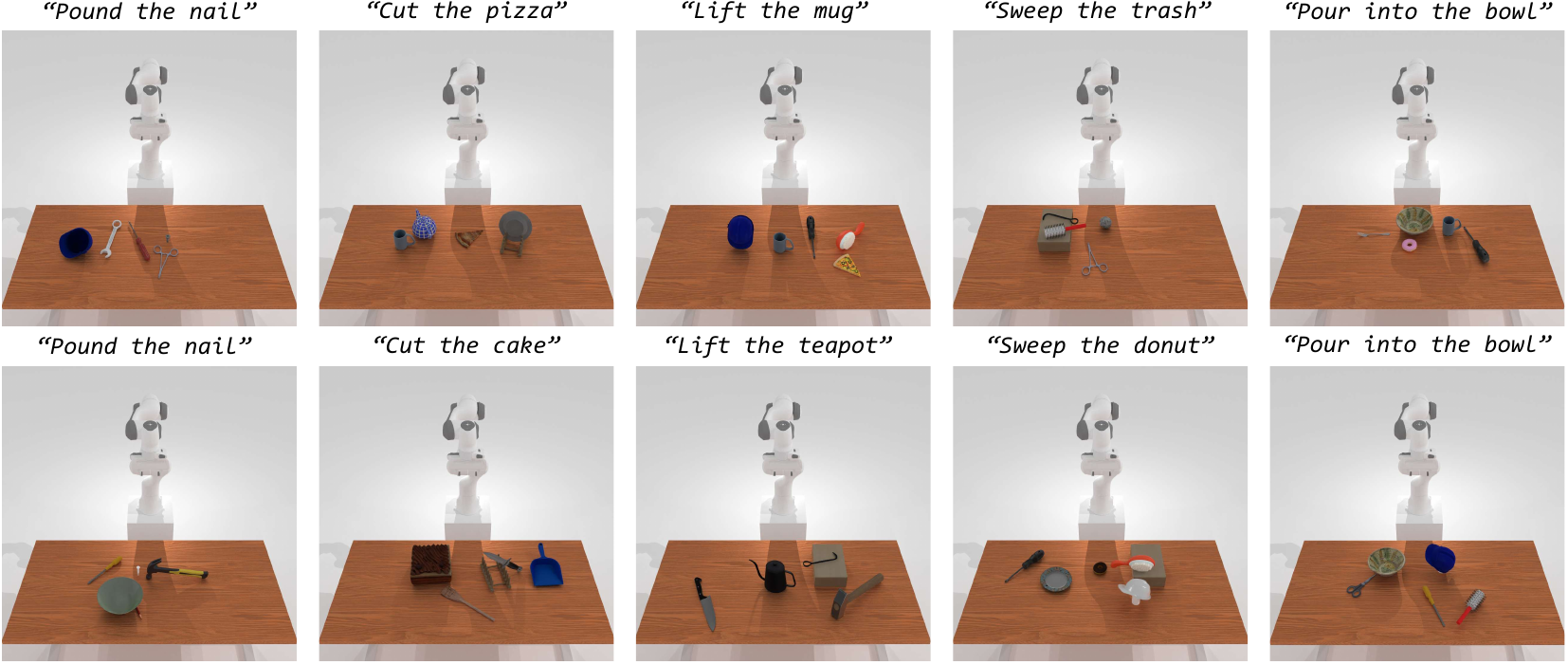}
  \caption{\textbf{Examples of tasks in simulation experiments.}}
  \label{fig:task}
\end{figure}

\begin{figure}[t]
  \centering
  \includegraphics[width=\linewidth]{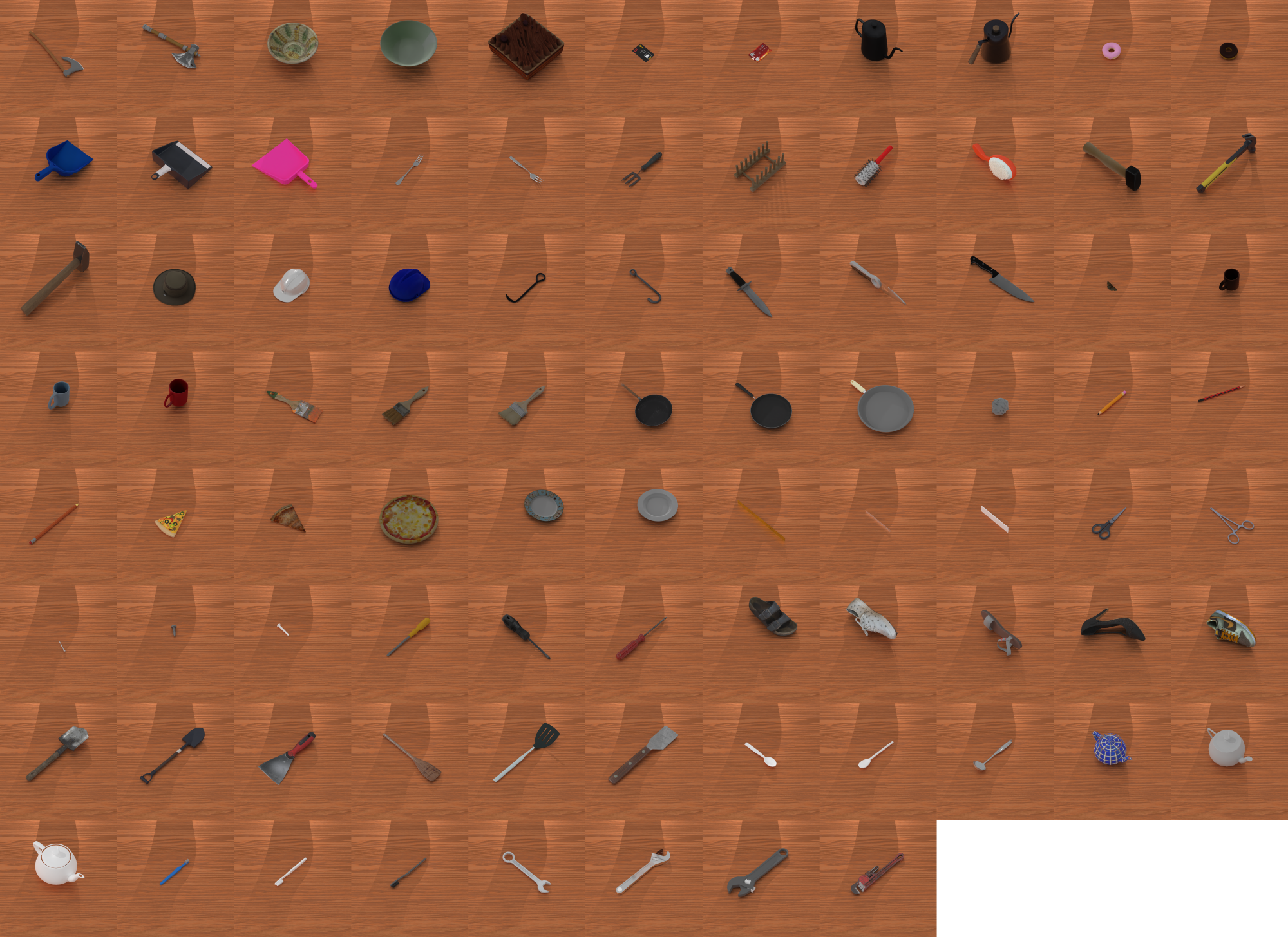}
  \caption{\textbf{Objects involved in the simulation experiments.}}
  \label{fig:objects}
\end{figure}

To evaluate whether the affordances predicted by \ours can effectively support robot tool use, we construct a tool-use benchmark in the SAPIEN 3 simulation environment~\cite{sapien} using a Franka Emika Panda arm. Each test case consists of a task instruction and a corresponding scene. To build diverse scenes for evaluation, we select 85 3D object assets from the Objaverse dataset~\cite{objaverse}, spanning a wide range of categories, sizes, shapes, and visual appearances, as shown in Fig.~\ref{fig:objects}. For each scene, we randomly sample objects based on the following rules: we first sample a target object mentioned in the task instruction, and then sample 3--5 additional objects as candidate tools. Each sampled object is placed on the table with a random position and orientation, while ensuring that it lies within the robot’s reachable workspace and does not collide with other objects. For evaluation, we ensure that at least one candidate tool in each scene can successfully complete the specified task. Using this procedure, we construct 100 diverse scenes. Fig.~\ref{fig:task} illustrates example scenes from our simulation experiments.

\begin{figure}[th]
    \centering
    \includegraphics[width=\linewidth]{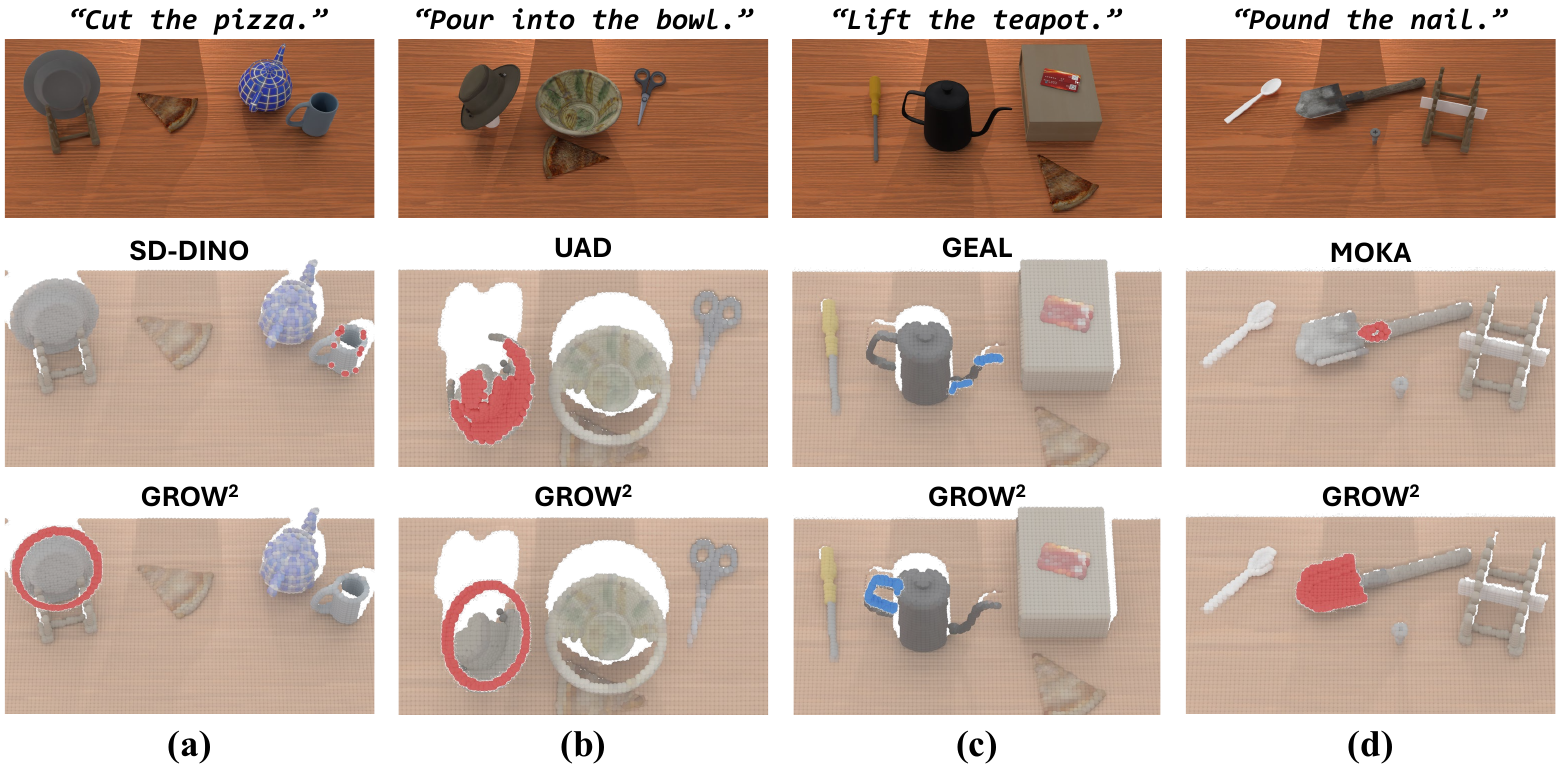}
    \caption{(a) SD-DINO~\cite{sd_dino} fails to select an appropriate tool; (b) UAD~\cite{uad} fails to accurately detect affordances under large-scale occlusion; (c) GEAL~\cite{geal} fails to generalize to novel object–function pairs; (d) MOKA~\cite{moka} fails to select the correct keypoint as the affordance. In (a), (b), and (d), \textcolor{red}{red} marks the affordance region on the tool object; in (c), \textcolor{blue}{blue} marks the affordance region on the target object.}
     \label{fig:baseline_failure}
\end{figure}

\subsubsection{Baselines}
We compare \ours against the same baselines as in \ourbench: DINOv3, SD-DINO, and MOKA. We also include UAD~\cite{uad} and GEAL~\cite{geal}, the best-performing methods on AGD20K and PIAD, respectively. Since UAD and GEAL are not designed for multi-object scenes, we provide them with the ground-truth tool. In addition, we evaluate ICP~\cite{ICP} as a one-shot 3D matching baseline.

For ICP, we use the complete 3D point clouds of all objects in the scene provided by the simulator. Given the demonstrated tool point cloud, we first identify the scene object whose point cloud best matches it. To improve matching robustness, we compute FPFH~\cite{rusu2009fast} descriptors for each pair of point clouds and use them to obtain an initial estimate of the 3D rigid transformation. Starting from this initialization, we then run Iterative Closest Point (ICP)~\cite{ICP} to refine the transformation and obtain the final alignment. This procedure produces an ICP matching score and a corresponding 3D rigid transformation for each candidate object in the scene. We select the object with the highest matching score as the tool. Finally, we transfer the demonstrated affordance region by applying the estimated 3D rigid transformation to it, yielding the corresponding affordance region on the selected tool.

\subsubsection{Metrics}
We define a success metric for each task type:

\xhdr{Pound}. A hard surface of the tool strikes the target object, delivering a large impact force. Success requires sufficient contact area and impact force.

\xhdr{Cut}. A sharp edge of the tool intersects the target object. Success requires the intersection region to be sufficiently sharp.

\xhdr{Lift}. A straight, thin part of the tool is inserted into the target object to lift it upward. Success requires sufficient lifting displacement.

\xhdr{Sweep}. A flat surface of the tool pushes the target object in a specified direction by a specified distance. Success requires the displacement error to be below a predefined threshold.

\xhdr{Pour}. A small object is placed in the selected tool after the robot reaches a pre-pouring pose. Success requires the object to fall into the target object after pouring.

\subsubsection{Qualitative Results}
Fig.~\ref{fig:baseline_failure} shows representative failure cases of baseline methods in comparison with \ours. Matching-based approaches such as SD-DINO~\cite{sd_dino} can fail to identify a functionally appropriate tool from a cluttered scene, since visual similarity alone does not necessarily reflect whether an object can be used to complete the task. In contrast, \ours selects tools in a symbolic space, allowing it to leverage the commonsense knowledge encoded in the VLM and reason about object functionality beyond appearance-level matching.

Existing methods also struggle with affordance localization. UAD~\cite{uad} fails to accurately detect affordances under severe occlusion, while GEAL~\cite{geal} has difficulty generalizing to novel object--function pairs. Although MOKA~\cite{moka} also uses a VLM to select the tool and target object, it relies on visual prompting to identify object keypoints as affordances, which makes its predictions less stable. By contrast, \ours avoids directly querying the VLM for fine-grained spatial localization and instead grounds the selected task-relevant parts through 3D reconstruction and multi-view segmentation, leading to more reliable affordance predictions. We attribute this advantage to the fact that VLMs are stronger at semantic reasoning than precise spatial reasoning; when asked to localize fine-grained regions directly, they are more prone to hallucination.

\end{document}